\documentclass[twocolumn]{IEEEtran}
\usepackage{multirow}
\usepackage{amsmath}
\usepackage{amsfonts}
\usepackage{subfigure}
\usepackage[utf8]{inputenc}
\usepackage[export]{adjustbox}
\usepackage{cite}
\usepackage{dirtytalk}

\graphicspath{{Images/}}

\DeclareMathOperator*{\maximize}{maximize}

\begin{document}

\title{Secure and Trustworthy Artificial Intelligence-Extended Reality (AI-XR) for Metaverses}
\author{Adnan Qayyum$^{1,2}$, Muhammad Atif Butt$^2$, Hassan Ali$^2$, Muhammad Usman$^3$, Osama Halabi$^4$, Ala Al-Fuqaha$^{5}$, \\ Qammer H. Abbasi$^1$, Muhammad Ali Imran$^1$ and Junaid Qadir$^{4*}$\thanks{\textit{$^*$Corresponding author: Junaid Qadir (jqadir@qu.edu.qa)}}\\
$^1$ James Watt School of Engineering, University of Glasgow, Glasgow, United Kingdom \\
$^2$ Information Technology University (ITU), Punjab, Lahore, Pakistan\\
$^3$ Glasgow Caledonian University, Glasgow, United Kingdom \\
$^4$ Qatar University, Doha, Qatar\\
$^5$ Hamad Bin Khalifa University, Doha, Qatar \\}

\maketitle

\begin{abstract}
Metaverse is expected to emerge as a new paradigm for the next-generation Internet, providing fully immersive and personalised experiences to socialize, work, and play in self-sustaining and hyper-spatio-temporal virtual world(s). The advancements in different technologies like augmented reality, virtual reality, extended reality (XR), artificial intelligence (AI), and 5G/6G communication will be the key enablers behind the realization of AI-XR metaverse applications. While AI itself has many potential applications in the aforementioned technologies (e.g., avatar generation, network optimization, etc.), ensuring the security of AI in critical applications like AI-XR metaverse applications is profoundly crucial to avoid undesirable actions that could undermine users' privacy and safety, consequently putting their lives in danger. To this end, we attempt to analyze the security, privacy, and trustworthiness aspects associated with the use of various AI techniques in AI-XR metaverse applications. Specifically, we discuss numerous such challenges and present a taxonomy of potential solutions that could be leveraged to develop secure, private, robust, and trustworthy AI-XR applications. To highlight the real implications of AI-associated adversarial threats, we designed a metaverse-specific case study and analyzed it through the adversarial lens. Finally, we elaborate upon various open issues that require further research interest from the community. 
\end{abstract}

\section{Introduction}
In the recent era, metaverse technology is rapidly emerging and there are a lot of potential applications that can benefit from these developments such as healthcare, industry, business, etc. While there is no single agreed-upon definition of a metaverse \cite{cheng2022will}, the metaverse is a convergence of physical, augmented, and virtual reality and provides a powerfully immersive experience to users by allowing them to seamlessly interact with the real and virtual (computer-simulated) world. The term ``metaverse'' is a combination of two terms, i.e., ``meta'' which means transcending, and ``universe'' which refers to the physical universe and the current virtual world. This is the basic definition of the term metaverse, nevertheless, the literature shows that it does not have a unified definition \cite{cheng2022will}.

Metaverse allows the creation of shared virtual space by connecting all virtual worlds through the Internet, where digital avatars (i.e., users) can communicate and interact with each other similar to the physical world. Key backbone technologies in the metaverse include artificial intelligence (AI) and extended reality (XR) that leverage different technological developments such as virtual reality (VR), augmented reality (AR), and mixed reality (MR). In addition, similar to the current Internet, metaverse will leverage other concomitant technologies like information and communication technologies (ICT), 5G, and 6G, but metaverse will provide a qualitatively different experience to its users by enabling real-life-like 3-D experiences through the incorporation of aforementioned technologies. 

Metaverse allows for the digitalization of traditional brick-and-mortar institutions and businesses—it will be possible to develop virtual markets, digital lands, and digital infrastructure, which can be bought and sold using blockchain and non-fungible tokens (NFTs), which are non-interchangeable units of data stored on a blockchain. Metaverse can be a game changer in terms of the impact of its potential applications due to the greater immersion, involvement, and personalization possible due to AI-XR. This is the prime reason various corporations have shown great interest in the idealization of the metaverse and are making big bets on developing their own AI-XR-based metaverse ecosystems.

\begin{table*}[!ht]
\centering
\caption{Comparison of our paper with existing surveys and review papers that are focused on analyzing privacy and security of AI-XR metaverse applications. (Legend: S $\rightarrow$ Security; P $\rightarrow$ Privacy; R $\rightarrow$ Regulatory; T $\rightarrow$ Trustworthy; $\surd$ $\rightarrow$ covered; $\times$ $\rightarrow$ Not Covered; $\approx$ $\rightarrow$ Partially Covered)}
\label{tab:sota_comp}
\scalebox{0.75}{
\begin{tabular}{|cc|p{100mm}|c|c|c|ccccc|c|c|}
\hline
\multicolumn{1}{|c|}{\multirow{2}{*}{\textbf{Year}}} & \multicolumn{1}{c|}{\multirow{2}{*}{\textbf{Authors}}} & \multirow{2}{*}{\textbf{Focused Area}} & \multicolumn{3}{c|}{\textbf{General Issues}} & \multicolumn{5}{c|}{\textbf{ML Related Issues \& Solutions}} & \multirow{2}{*}{\textbf{\begin{tabular}[c]{@{}c@{}}Background \& \\ Applications\end{tabular}}} & \multirow{2}{*}{\textbf{\begin{tabular}[c]{@{}c@{}}Open \\ Issues\end{tabular}}} \\ \cline{4-11}
\multicolumn{1}{|c|}{} & \multicolumn{1}{c|}{}  &  & \multicolumn{1}{c|}{\textbf{S}} & \multicolumn{1}{c|}{\textbf{P}} & \textbf{R} & \multicolumn{1}{c|}{\textbf{S}} & \multicolumn{1}{c|}{\textbf{P}} & \multicolumn{1}{c|}{\textbf{T}} & \multicolumn{1}{l|}{\textbf{XAI}} & \multicolumn{1}{l|}{\textbf{Ethical}} &  &  \\ \hline

\multicolumn{1}{|c|}{2018} & \multicolumn{1}{c|}{Falchuk et al. \cite{falchuk2018social}} & Privacy issues and solutions for digital footprints in metaverse games. & $\surd$ & $\surd$ & $\times$ & \multicolumn{1}{c|}{$\times$} & \multicolumn{1}{c|}{$\times$} & \multicolumn{1}{c|}{$\times$} & \multicolumn{1}{c|}{$\times$} & $\times$ & $\approx$ & $\times$  \\ \hline

\multicolumn{1}{|c|}{2020} & \multicolumn{1}{c|}{Guzman et al. \cite{de2019security}} & Analyzed privacy and security in MR from data-centric perspective. & $\surd$ & $\surd$ & $\approx$  & \multicolumn{1}{c|}{$\times$} & \multicolumn{1}{c|}{$\times$} & \multicolumn{1}{c|}{$\times$} & \multicolumn{1}{c|}{$\times$} & $\times$ & $\approx$ & $\surd$  \\ \hline

\multicolumn{1}{|c|}{2021} & \multicolumn{1}{c|}{Ning et al. \cite{ning2021survey}} & General focused survey on metaverse with partial discussion on privacy and security issues. & $\approx$ & $\approx$ & $\times$ & \multicolumn{1}{c|}{$\times$} & \multicolumn{1}{c|}{$\times$} & \multicolumn{1}{c|}{$\times$} & \multicolumn{1}{c|}{$\times$} & $\times$ & $\surd$ & $\surd$  \\ \hline

\multicolumn{1}{|c|}{2021} & \multicolumn{1}{c|}{Pietro et al. \cite{dimetaverse}} & Discussed general privacy and security issues in metaverse applications. & $\surd$ & $\surd$ & $\times$ & \multicolumn{1}{c|}{$\times$} & \multicolumn{1}{c|}{$\times$} & \multicolumn{1}{c|}{$\times$} & \multicolumn{1}{c|}{$\times$} & $\times$ & $\approx$ & $\surd$  \\ \hline

\multicolumn{1}{|c|}{2022} & \multicolumn{1}{c|}{Huynh et al. \cite{huynh2022artificial}} & Discussed potential applications of AI in various metaverse applications. & $\approx$ & $\approx$ & $\times$  & \multicolumn{1}{c|}{$\times$} & \multicolumn{1}{c|}{$\times$} & \multicolumn{1}{c|}{$\times$} & \multicolumn{1}{c|}{$\times$} & $\times$ & $\surd$ &$\approx$  \\ \hline

\multicolumn{1}{|c|}{2022} & \multicolumn{1}{c|}{Zhao et al. \cite{zhao2022metaverse}} & Security \& privacy issues and solutions for four dimensions: communication, user information, scenario, and goods. & $\surd$ & $\surd$ & $\times$  & \multicolumn{1}{c|}{$\times$} & \multicolumn{1}{c|}{$\times$} & \multicolumn{1}{c|}{$\times$} & \multicolumn{1}{c|}{$\times$} & $\times$ & $\approx$  & $\times$   \\ \hline

\multicolumn{1}{|c|}{2022} & \multicolumn{1}{c|}{Wang et al. \cite{wang2022survey}} & Presented general (non ML-associated) security and privacy related challenges for different metaverse applications. & $\surd$ & $\surd$ & $\surd$  & \multicolumn{1}{c|}{$\times$} & \multicolumn{1}{c|}{$\times$} & \multicolumn{1}{c|}{$\times$} & \multicolumn{1}{c|}{$\times$} & $\times$ & $\surd$  & $\surd$  \\ \hline

\multicolumn{2}{|c|}{This Paper} & ML-associated security, privacy, and trustworthiness challenges and solutions for AI-XR metaverse applications. & $\surd$ & $\surd$ & $\surd$ & \multicolumn{1}{c|}{\begin{tabular}[c]{@{}c@{}}  $\surd$ \end{tabular}} & \multicolumn{1}{c|}{\begin{tabular}[c]{@{}c@{}} $\surd$ \end{tabular}} & \multicolumn{1}{c|}{\begin{tabular}[c]{@{}c@{}}  $\surd$ \end{tabular}} & \multicolumn{1}{c|}{\begin{tabular}[c]{@{}c@{}}  $\surd$ \end{tabular}} & \begin{tabular}[c]{@{}c@{}}  $\surd$ \end{tabular} & $\surd$  & $\surd$  \\ \hline
\end{tabular}}
\end{table*}

Metaverse is receiving increasing traction from numerous major tech companies worldwide such as Facebook (which is recently rebranded with the name ``Meta''), Microsoft, Google, and Amazon. In addition, the widespread adoption of the metaverse is evident in the infusion of billions of dollars of investments by these companies in an attempt to achieve great technological transformation. However, despite such huge traction of the metaverse and its potential to transform existing ecosystems like healthcare, there are numerous challenges associated with the use of AI in the metaverse that may hinder their seamless adoption by end users in the longer term. In addition, in the backdrop of recent technologically induced social issues, there is a palpable lack of trust and confidence in such technologies. 

Since technology can be used both ways (i.e., for good and harm), it is vital that governments, corporations, and society at large seriously consider ethical and moral issues. There are many ethical questions about privacy, security, transparency, accountability, democracy, freedom of speech, and anonymity that technology alone cannot answer. Some specific concerns related to how AI-XR-based metaverse applications will impact humanity are: (1) how AI-XR-based metaverse applications will impact and promote human values and human rights? how will AI-XR-based metaverse promote social welfare and not cause harm to society at large; (3) how can we regulate critical applications of metaverse like healthcare? (4) How do we align the commercial and technical imperatives of AI-XR metaverse applications with human values and promote a moral vision and character development? (5) How do we ensure that AI-XR metaverse application developers do not exploit or manipulate their users? Keeping in mind the aforementioned questions, in this paper, we present an analysis of the security, privacy, and trust issues associated with the use of AI-XR in metaverse applications. 

\textit{Contributions of this Paper:} To the best of our knowledge this paper is the first attempt towards analyzing the challenges associated with the use of different AI techniques in AI-XR metaverse applications. The comparison of this paper with existing survey and review articles that are focused on analyzing security and privacy aspects of AI-XR metaverse applications is presented in Table \ref{tab:sota_comp}. In the summary, the following are the salient contributions of this paper.

\begin{enumerate}
    \item We highlight various issues that arise with the use of AI in metaverse applications that mainly include security, privacy, and trustworthiness. 
    \item We present a taxonomy of different potential solutions that can be used to realize secure, robust, safe, and trustworthy AI-XR applications. 
    \item We identify various ML-based use cases across different layers of metaverse architecture and highlight several ML-associated vulnerabilities at each layer.  
    \item We present a case study to highlight the real threat of AI-based vulnerabilities by considering a prospective metaverse application design scenario.  
    \item We elaborate upon various open issues that require further development. 
\end{enumerate}

\begin{figure*}[!ht]
    \centering
    \includegraphics[width=0.75\textwidth]{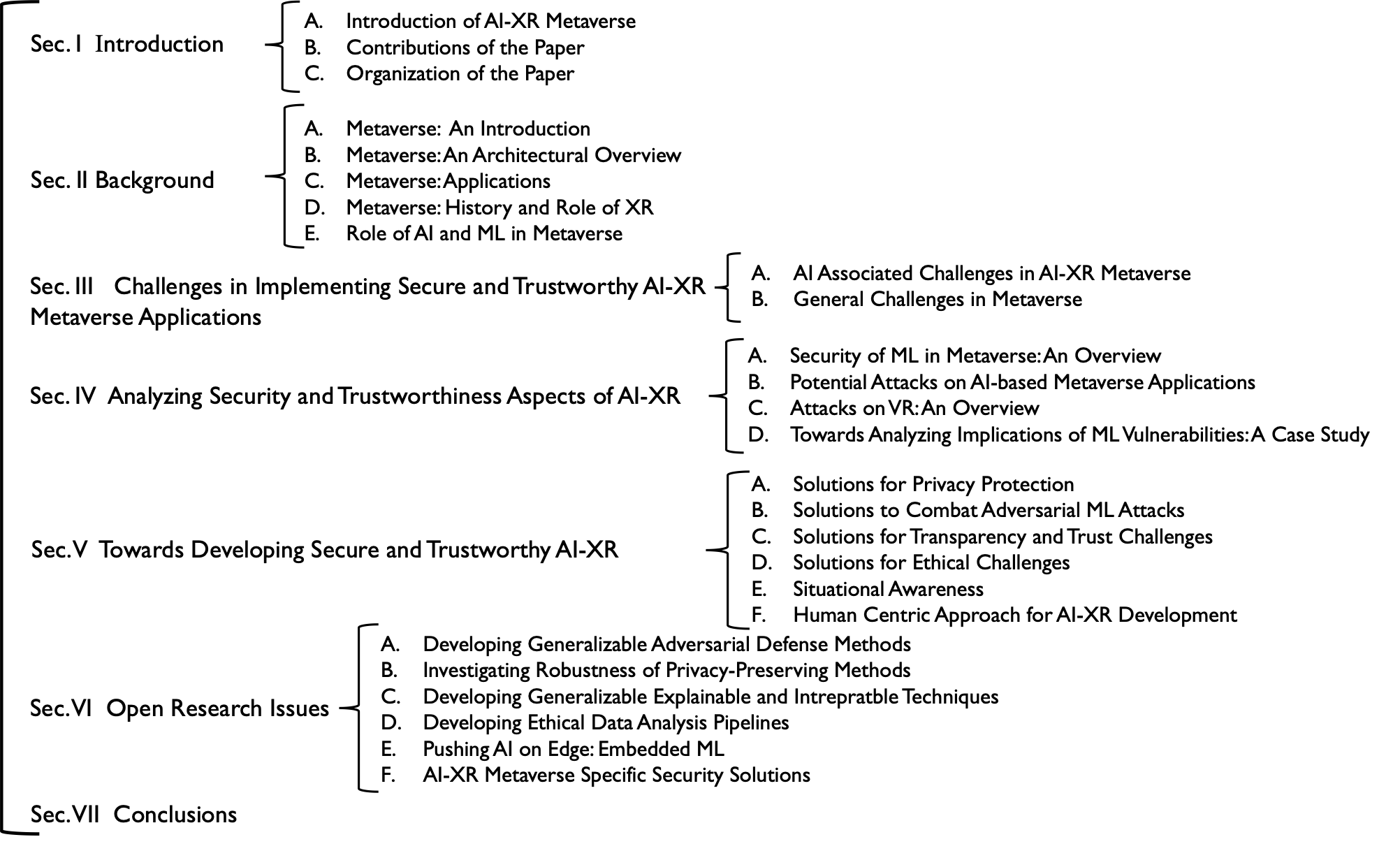} 
    \caption{Organization of the paper. }
    \label{fig:organization}
\end{figure*}

\textit{Organization of this Paper:} The rest of the paper is organized as follows. Section \ref{sec:back} presents relevant background. The discussion of challenges related to security, safety, privacy, and trust is presented in Section \ref{sec:challenges}. The taxonomy of different vulnerabilities associated with the use of ML in AI-XR metaverse applications is discussed in Section \ref{sec:SecML}. Different potential solutions that can ensure security, privacy, safety, and trust in ML applications are discussed in Section \ref{sec:solutions}. Various open issues that require further research attention are highlighted in Section \ref{sec:open}. Finally, we conclude the paper in Section \ref{sec:con}. The organization of the paper is depicted in Figure \ref{fig:organization}. 

\begin{figure*}[!t]
    \centering
    \includegraphics[width=\textwidth]{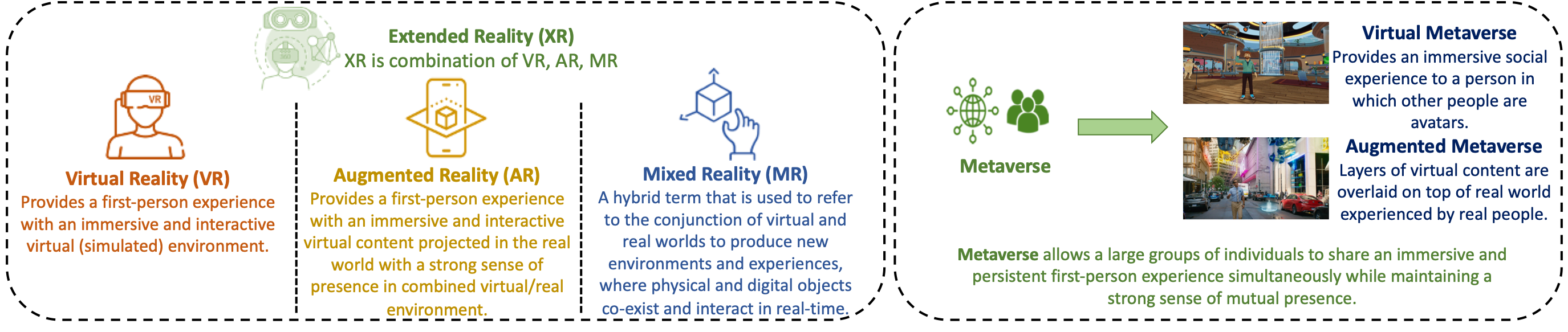} 
    \caption{An overview of different concepts related to metaverse that include VR, AR, MR, XR, virtual metaverse, and augmented metaverse. }
    \label{fig:metaverse_overview}
\end{figure*}


\section{Background}
\label{sec:back}

\subsection{Metaverse: An Introduction}
Before understanding the concept of the metaverse, it is very important to understand the related concepts that are described below.

\begin{itemize}
    \item \textit{Virtual Reality (VR):} In VR, the users achieve an immersive experience by donning a VR headset that allows them to enter into a virtual (computer-simulated) world thus completely blanking out the real world. The key objective of enabling immersion in VR is to provide high fidelity user interaction to give him the feeling that the virtual world is real \cite{Halabi2020}. Prime examples of VR include Facebook Oculus and HTC VIVE VR headrests. VR has a wide range of applications but a VR headset is required to enter the digital world.   
    \item \textit{Augmented Reality (AR):} In AR, the users obtain an immersive experience by blending the virtual (digital) and the real world and projecting digital content (text, images, and sounds) onto the real world. Unlike VR, AR can be realized without special equipment (like a headset) through the use of smartphones, implants, glasses, or contact lenses that are used to overlay digital content on top of the real world. 
    \item \textit{Mixed Reality (MR):} MR is a hybrid term that is used to refer to the conjunction of virtual and real worlds to produce new environments and experiences, where physical and digital objects co-exist and interact in real-time (it is an enhanced form of AR). Microsoft HoloLens headrest is an example MR headset.
    \item \textit{Extended Reality (XR):} XR is an umbrella term that encompasses and subsumes VR, AR, and MR. It covers all the future realities that might emerge from these technologies. XR is predicted to become a \$209 billion market by 2022.
\end{itemize}

\begin{figure*}[!t]
    \centerline{\includegraphics[width=0.98\textwidth]{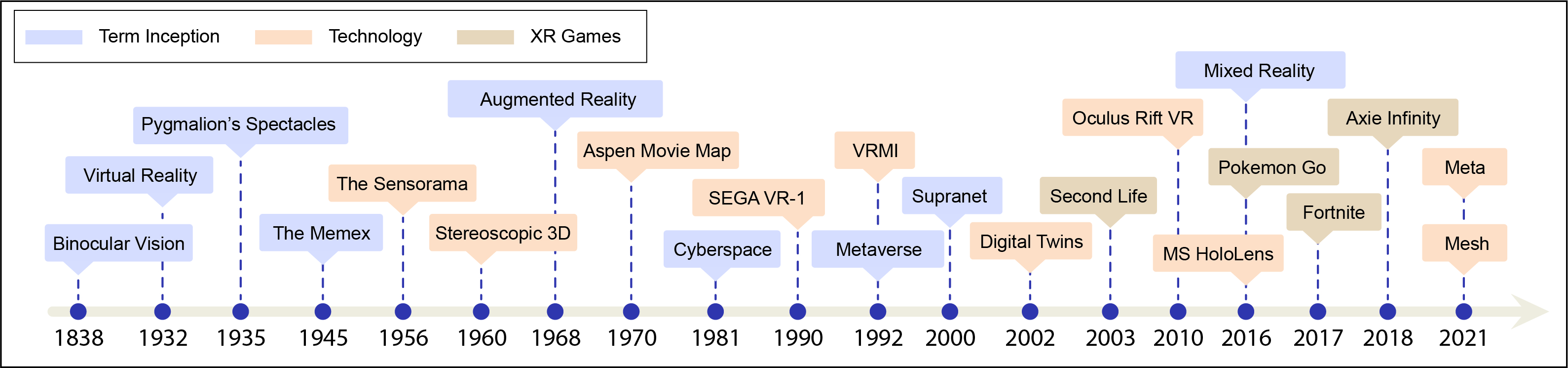}}
    \caption{Historical evolution of technologies needed for metaverse}
    \label{fig:BG-Metaverse}
\end{figure*}

\begin{figure}[!t]
    \centering
    \includegraphics[width=0.45\textwidth]{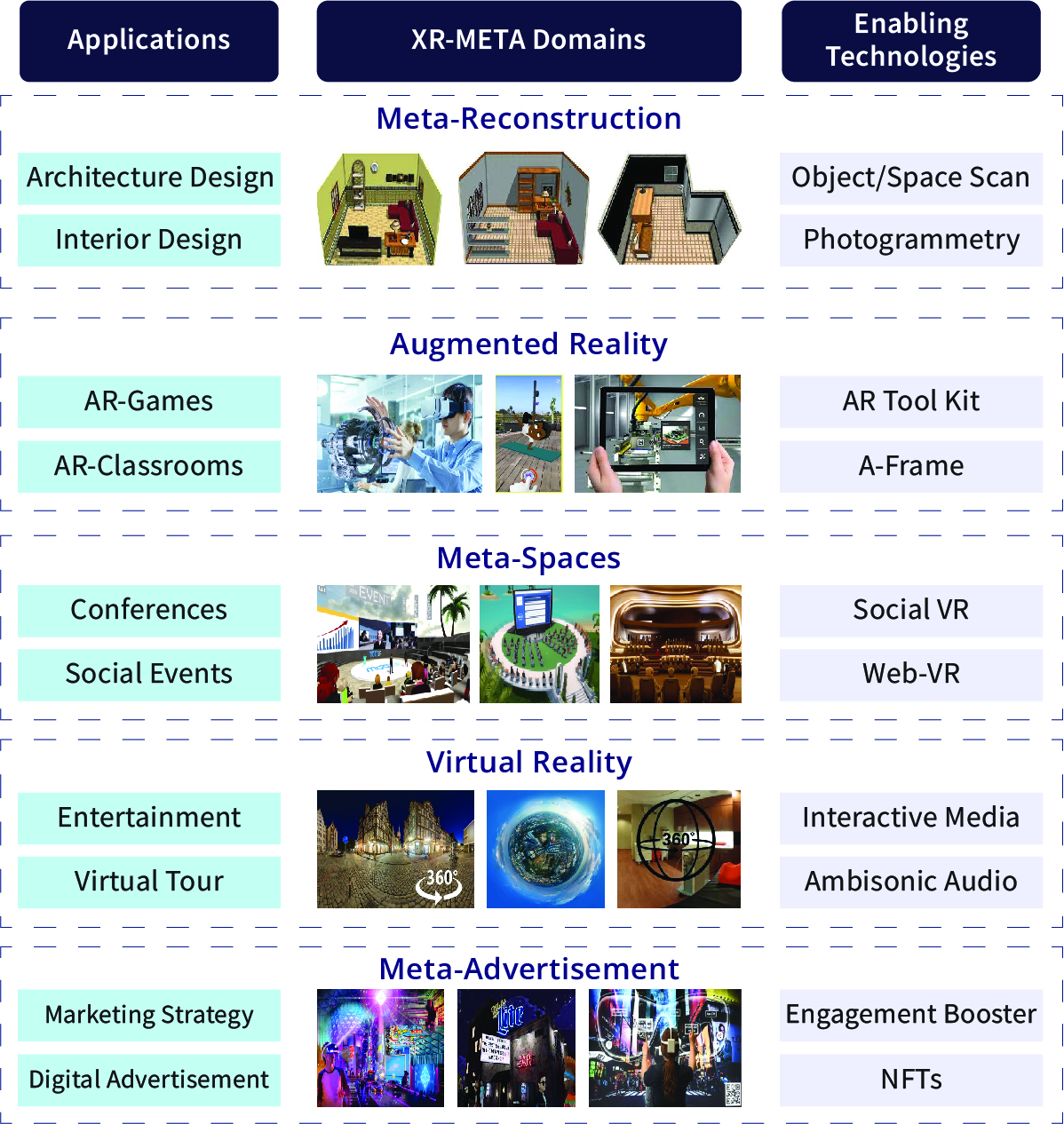} 
    \caption{Applications of XR in metaverse.}
    \label{fig:XR_meta_apps}
\end{figure}

Immersive first-person experiences are one of the most significant aspects of XR, VR, and AR. The Metaverse takes this to the next level, allowing large groups of individuals to share an immersive first-person experience while maintaining a strong sense of mutual presence. Although the term ``metaverse" is often associated with virtual reality, according to Rosenberg, there are two types of metaverses: a ``virtual metaverse" in which people are avatars and an ``augmented metaverse" in which layers of virtual content are overlaid on the real world and experienced by real people. Figure \ref{fig:metaverse_overview} depicts XR, VR, AR, and the virtual and augmented metaverse, as well as their interaction. Metaverse is the next generation of the Internet that will surround us both graphically and socially. A historical overview of developments regarding metaverse is shown in Figure \ref{fig:BG-Metaverse} and different applications of XR in metaverse along with their enabling technologies are illustrated in Figure \ref{fig:XR_meta_apps}. 


\subsection{Metaverse: An Architectural Overview}

\begin{figure}[!t]
    \centering
    \includegraphics[width=0.45\textwidth]{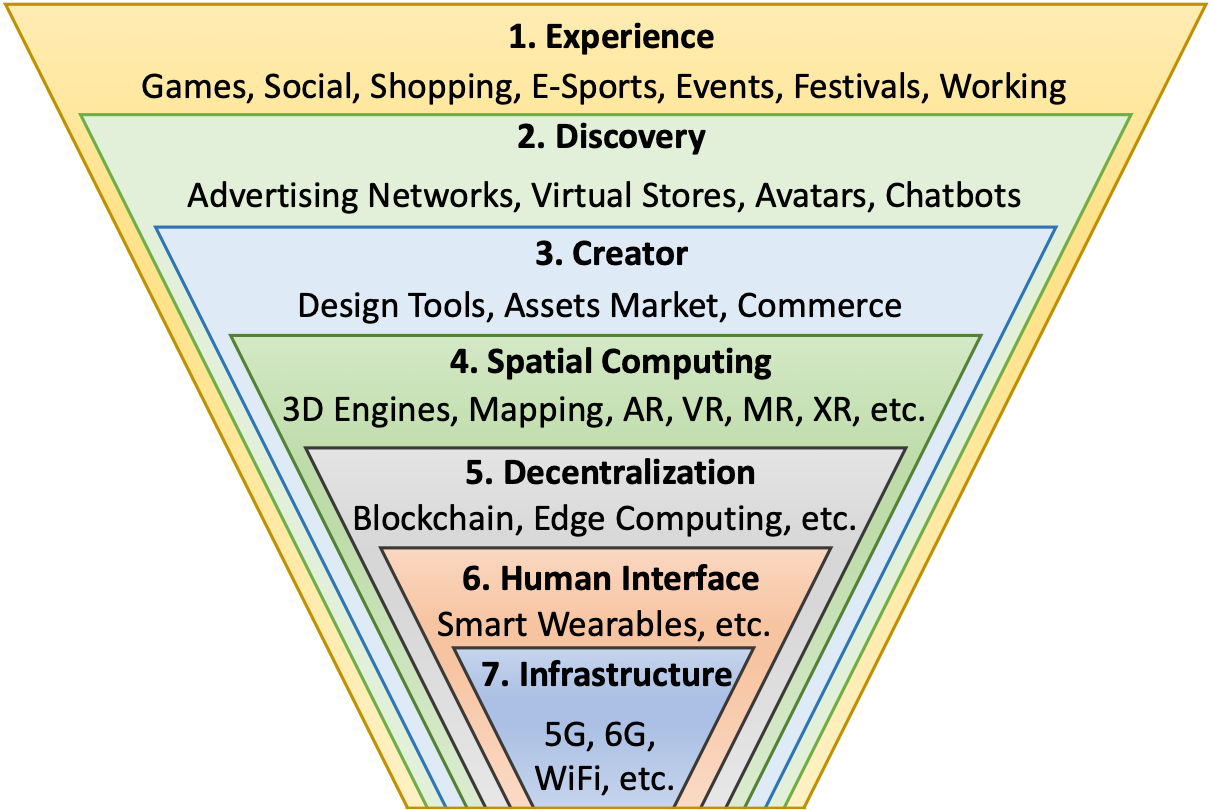} 
    \caption{Illustration of different layers in metaverse.}
    \label{fig:metaverse_arch}
\end{figure}

The architecture of the metaverse expands from the people's experiences to the underlying enabling technologies and has seven layers that include: (1) Experience; (2) Discovery; (3) Creator; (4) Spatial Mapping; (5) Decentralization; (6) Human Interface; and (7) Infrastructure, which is illustrated in Figure \ref{fig:metaverse_arch} and is briefly described below. 

\begin{itemize}
    \item[-] \textit{Layer 1: Experience:} It is the topmost layer in the metaverse, which is mainly concerned with the experiences of the users and it provides different services to them, e.g., games, E-sports, social interactions, events, festivals, shopping, co-working, etc.
    \item[-] \textit{Layer 2: Discovery:} It is like a push and pull service that introduces people to new experiences in the metaverse such as virtual stores, advertising networks, ratings, social curation avatars, chatbots, etc. It will involve both inbound (i.e., users are actively seeking information regarding experiences) and outbound (i.e., an advertisement that is not explicitly requested by the user). This layer is mainly driven by metaverse service providers and content creators to inform and motivate users regarding new features and services. 
    \item[-] \textit{Layer 3: Creator:} This layer is sometimes also referred to as the creator economy. Like the previous layer, it is mainly driven by the content creator and service providers, who leverage different technologies to create content or experiences for metaverse users such as asset markets, E-commerce, design tools, and workflow. 
    \item[-] \textit{Layer 4: Spatial Mapping:}  This layer provides a bridge between the digital world and the physical world and provides immersive experiences to metaverse users. It consists of different technologies like geospatial mapping, object and speech recognition, 3D engines (for enabling animations), VR, AR, XR, multitasking, and integration of user interfaces and heterogeneous sensor data (e.g., from IoT and wearable devices). It can be assumed as the backbone of the creator layer \cite{far2022applying}. 
    \item[-] \textit{Layer 5: Decentralization:} Decentralization is very crucial in the metaverse and ideally it should not be controlled by a single entity. It provides a scalable ecosystem to developers in terms of providing online capabilities and reliability to the users at the same time. This layer will consist of multiple technologies that include edge computing, blockchain, microservices, and AI agents. 
    \item[-] \textit{Layer 6: Human Interface:} This layer is mainly concerned with the interfacing of the physical world with the digital and from the digital to the physical world. For example, let's consider the example of metaverse services that require data collected from humans using different sensors such as smartwatches, smartphones, smart glasses, wearable IoT devices, biosensors, and head-mounted displays, just to mention a few. 
    \item[-] \textit{Layer 7: Infrastructure:} This layer is responsible for connecting different enabling devices and technologies to the network for content delivery in the metaverse. Different ICT technologies will act as a backbone in the infrastructure layer of the metaverse. For example, 5G/6G-based communication has huge potential to drastically improve the performance of metaverse applications while reducing latency and speeding up content delivery. In addition, this layer will also involve major data processing capabilities like data centers, the cloud, CPUs, GPUs, and even quantum computers.  
    
\end{itemize}

\begin{figure*}[!t]
    \centering
    \includegraphics[width=0.93\textwidth]{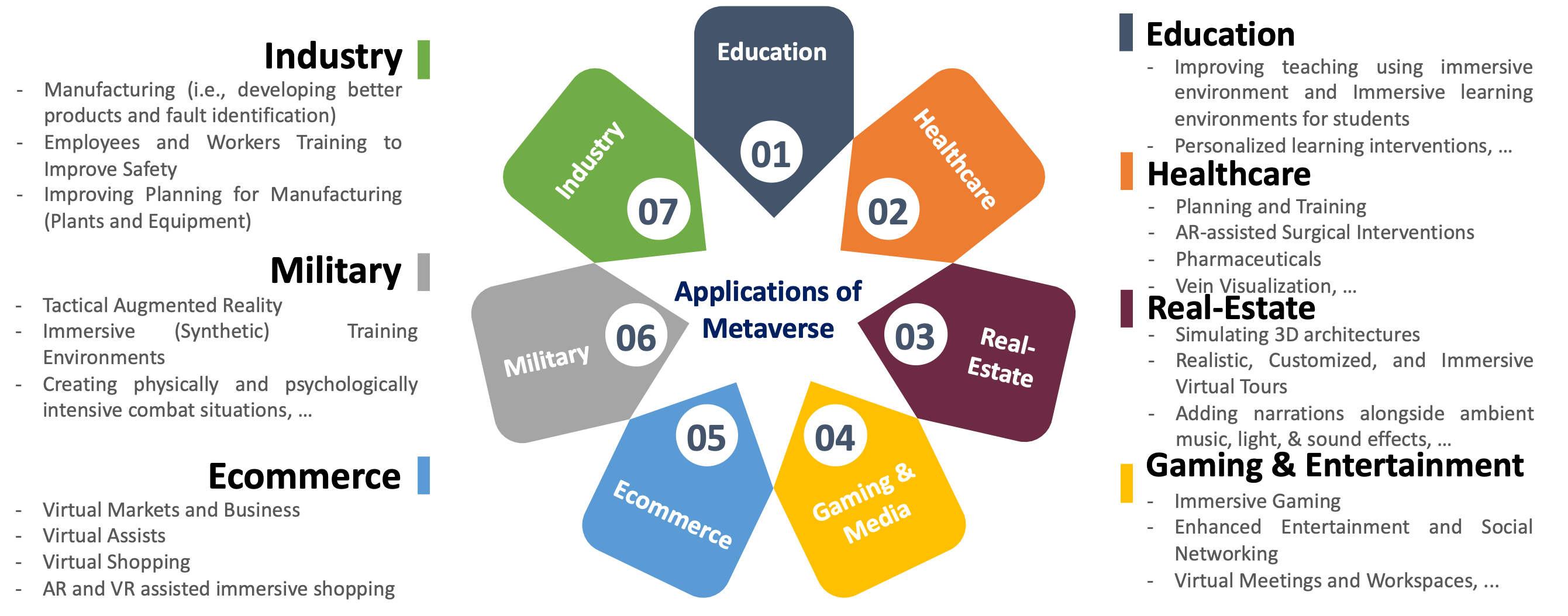} 
    \caption{Various potential AI-XR metaverse applications.}
    \label{fig:meta_apps}
\end{figure*}

\subsection{Metaverse: Applications}
Metaverse applications that incorporate different technologies like VR, AR, or XR have various potential applications in education, healthcare, industry, and scientific research, just to name a few. A detailed taxonomy of various potential metaverse applications is illustrated in Figure \ref{fig:meta_apps}. Metaverse allows moving from text-focused Internet that supports 2D images to a 3D or even a 4D world (in which we may travel in time (forward or backward)). One of the promising applications will be social VR, which will be the enhanced version of current social media. As metaverse can leverage both VR and AR, numerous applications (e.g., voice recognition, gesture recognition, and speech translation) can benefit \cite{reiners2021combination}. Over the past few years, VR and AR technologies have become very mature and nowadays their equipment is relatively cheap and readily available. This is a long way from the modest beginning of AR and VR, which were ignited in the 1960s by Ivan Sutherland in his pioneering work on the first responsive head-mounted wearable devices, which were admittedly primitive by modern standards.

Modern VR headsets have become accessible (e.g., Facebook's Oculus Quest 2 is available for ~\$300) with the price expected to go down as technology continues to advance.  There are numerous AR applications such as Heads-Up-Display (HUD) features on modern luxury cars, the use of face filters in apps such as Snapchat, and games such as the addictive Pokémon GO game, where players could ``see'' Pokémon characters on the street. Modern mobile phones supporting Lidar technology now can support AR with new software development kits emerging such as Google's AR development platform ARCore, which provides the ability to track motion, understand the environment, and estimate light—three capabilities essential for AR. AR pioneer Louis Rosenberg predicts that within 10 years, most people will be spending more than 2 hours every day in VR, and augmented reality interfaces will replace mobile phones as our primary interface with digital content.

\begin{figure}[!t]
    \centering
    \includegraphics[width=0.48\textwidth]{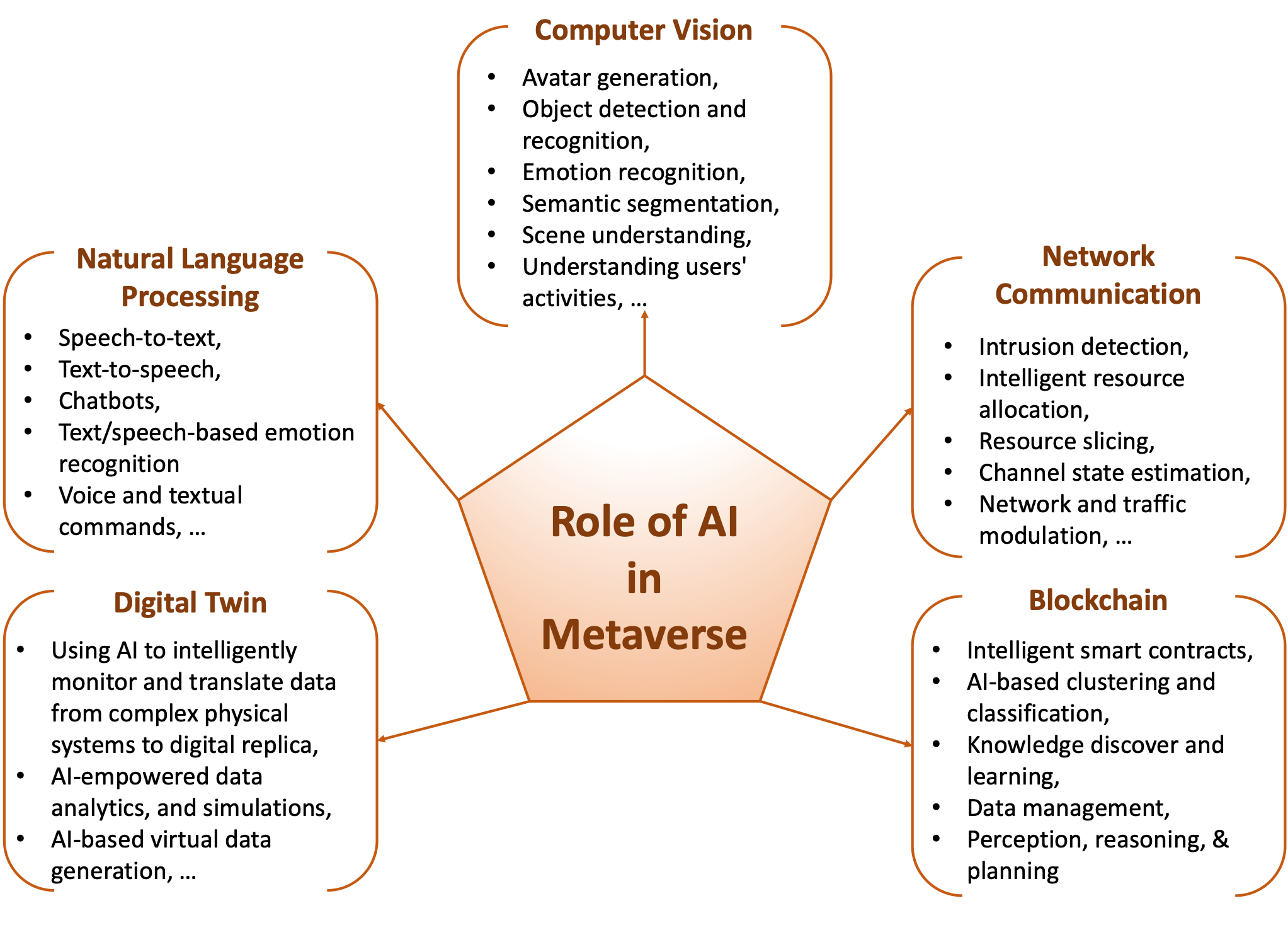} 
    \caption{Applications of AI in metaverse.}
    \label{fig:ML_meta_apps}
\end{figure}

\subsection{Role of AI and ML in Metaverse}

Different AI techniques including machine learning (ML) and deep learning (DL) have many potential applications in the metaverse (as shown in Figure \ref{fig:ML_meta_apps}). For example, one of the most fascinating features of the metaverse will be voice-based commands that will utilize different voice recognition and analysis techniques for language processing and understanding human commands. In addition, metaverse will use different ML/DL-based regression and classification for data management and decision making. Also, to provide immersive experiences to the users it will use different generative models to generate photo-realistic avatars and for 3D reconstruction of objects from 2D images. We discuss the potential applications of ML/DL in the metaverse across five dimensions, which are described next.

\subsubsection{Applications of ML in Natural Language Processing}
Natural Language Processing (NLP) consists of different techniques that are used for automatically analyzing and understanding human languages (i.e., text and speech). There are many NLP applications that will be part of AI-XR metaverse applications, e.g., speech-to-text, text-to-speech, chatbots, and text/speech-based emotion recognition are the most prominent features of the metaverse. In particular, NLP techniques will be used for the recognition and understanding of complicated human conversations and commands. A key driving force behind the success of NLP methods is the advancement in ML/DL, with the development of new techniques such as recurrent neural networks (RNN), long-short term memory (LSTM), and transformer networks  \cite{otter2020survey}.  

\subsubsection{Applications of ML in Vision}
Machine vision or computer vision will be a fundamental component of AI-XR metaverse applications. Different computer vision applications will enable various functionalities in metaverse applications, for example, processing visual data from different sensors to infer high-level visual semantics. The major tasks of the visual processing pipeline include understanding users' activities, emotion recognition, object detection, scene understanding, semantic segmentation, avatar generation, etc. In addition to these applications, the metaverse is expected to have AI-empowered quality assessment capabilities, e.g., for satisfying the users' demands about viewing high-resolution videos \cite{huynh2022artificial}. In this regard, advanced AI methods can be used to develop quantitative and qualitative benchmarks for visual quality assessment.  

\subsubsection{Applications of ML in Network Communication}
Metaverse is expected to simultaneously entertain a massive number of users with the metaverse services provisioned mainly through wireless networks. Over the past few years, substantial research attention has been devoted to improving the overall throughput and performance of wireless network communication and the use of different AI techniques is the main driving force behind this innovation \cite{chen2019artificial}. Metaverse will mainly include real-time multimedia services that require a reliable connection, high throughput, and low latency to ensure a seamless user experience. Therefore, it is expected that the metaverse will benefit from 5G and beyond empowered communication. The potential of different AI techniques has already been demonstrated for 5G and 6G, e.g., intelligent resource allocation \cite{she2020deep}, solving resource slicing problem \cite{alsenwi2021intelligent}, channel state estimation \cite{luo2018channel}, and network modulation \cite{tunze2020sparsely}, etc. 

\subsubsection{Applications of ML in Blockchain}
Service providers in the metaverse will provide users with different incentives in terms of digital assists (e.g., coins) for different events, games, and creative activities. The dispersion of such assets requires a transparent way to record and track such transactions. In this regard, smart contracts empowered blockchain technology can be leveraged that allows critical information to be stored on an immutable and impenetrable ledger. The decentralized nature of blockchain imbues it with great potential to address security and privacy issues in metaverse \cite{cannavo2020blockchain}. This potential increases with AI-empowered blockchain applications \cite{yang2022fusing}, for example, different AI-based clustering and classification techniques can be used for data analysis stored on blockchain \cite{tanwar2019machine}. In addition, different AI techniques can be used for knowledge discovery and learning, efficient data management, perception, reasoning, and planning. 

\subsubsection{Applications of ML in Digital Twin}
The term digital twin refers to the digital replica (i.e., representation) of real objects. A digital twin is capable of synchronizing regular actions, operations processes, and assets with the real world, e.g., analyzing, monitoring, predicting, and visualizing \cite{tao2018digital}. The digital twin also acts as a bridge where the actual world and digital world interact with each other through different IoT devices \cite{chen2021digital}. The digital twin will be one of the most important building sectors of the metaverse that allows users to access and use services in the virtual world while exactly depicting the real world in a virtual environment. For example, surgeons and medical experts can create a digital replica of a patient to study and understand the involved complexities before performing his surgery. 

\section{Challenges in Implementing Secure and Trustworthy AI-XR Metaverse Applications}
\label{sec:challenges}
Despite the significant potential of different AI-XR metaverse applications, there are various challenges related to security, privacy, and lack of trust that can hinder their wide adoption. A few such challenges include privacy breaches, security invasion, user profiling, unfair AI outcomes, etc. These challenges may directly or indirectly put the users' safety at risk and can influence social acceptability \cite{lee2021all}. Moreover, as discussed above metaverse is the integration of different modern technologies like AI, blockchain, and 5G/6G, therefore, it is likely that the inherent issues associated with these technologies get translated into the metaverse. In this section, we describe different challenges that can hinder the secure, safe, robust, and trustworthy employment of AI-XR metaverse applications. Specifically, we characterized and discuss these challenges in two dimensions, i.e., challenges associated with the use of AI techniques including ML/DL-based methods, and XR-related challenges in the metaverse. We will start by first discussing AI-related challenges. 


\subsection{AI Associated Challenges in AI-XR Metaverse}
Modern AI techniques that include ML/DL-based models suffer from different vulnerabilities that hinder the smooth, safe, secure, and trustworthy use of these methods in critical applications like healthcare, autonomous vehicles, and AI-XR metaverse applications. Below we briefly discuss various such challenges.  

\subsubsection{Privacy Issues}
Ensuring the privacy of the end users will be a major challenge in AI-XR metaverse applications. As these applications are designed to monitor and collect users' data at an unprecedented fine-grained level, in a bid to create a replica of the digital world \cite{falchuk2018social}, there is a greater danger and risk of privacy breaches \cite{wang2022survey}. For example, to create an immersive virtual scene in the metaverse, data from different sensors will be collected and analyzed using AI models, e.g., facial expressions, brain wave patterns, hand movements, eye movements, biometric, and speech data \cite{wang2022survey}. This raises obvious concerns regarding the privacy of users and opens a new horizon for digital crimes \cite{falchuk2018social}. Users' sensitive information including daily routine activities, personal logs, and schedules will be stored on a server, which ultimately becomes a critical privacy challenge in a publicly distributed network. Such data include body movements, voice, reflexes, and even more critical data that include subconscious and unconscious responses such as eye movements and physiological signals. Features such as eye tracking are readily accessible using commercially available products such as the HTC Vive Pro Eye and Pico Neo VR headsets even though the XR expert Louis Rosenberg recommends banning such features and data collection in non-health-related applications for ethical reasons. In general, the data is collected through on-device sensors at the user site, processed at their devices or nearby local server, and logged as storage in the cloud. Considering the above-mentioned procedure, some malicious attacks can be encountered which are classified as (i) data collection, (ii) data storage, (iii) data usage, and (iv) user profiling. Furthermore, as AR technology depends on the precise localization of users in the physical world, modern smartphones use Lidar sensors and Simultaneous Localization and Mapping (SLAM) algorithms to precisely locate the user in the real world. This opens up the possibility of privacy violations as sensitive information may be exploited for anti-social purposes. Some important concerns related to the privacy of data are described next.




\paragraph{Data Collection} 
Generally, the data in AI-XR metaverse applications is collected through users' input either by voice command or textual input, and multi-modal sensors including camera, microphone, textual commands, gesture sensors, and wearable devices. These devices are expected to frequently collect personal information such as daily routine activities, voice and biometric data, and personal preferences including shopping items, TV shows, and preferred food choices. In addition, the private data will be used for the creation of avatars for a digital representation of a real human in the metaverse, which also raises privacy issues. For example, the built-in location sensor in the Oculus headset can be used for tracking users' presence in the real environment with a precise accuracy \cite{wang2022survey}.   

\paragraph{Data Storage} 
The ultimate aim of developing personalized AI-XR metaverse applications is to aid human beings and ease their daily routine activities. These applications will contain multiple sensing devices which generate a substantial amount of data. Whereas these devices will be resource constrained with small storage units, which leads to the tradeoff between data generation and storage at the edge level. To address these shortcomings, these devices upload their data along with the corresponding logs to online local or global data centers. Though, the data storage tradeoff is resolved by connecting with the servers. However, it also raises privacy concerns regarding the access permissions and data protection of the consumers. Also, if the communication channel is hacked by an attacker then the data can be manipulated to get the intended outcomes. The literature suggests that adversaries can extract information regarding the actual data even if the communication is encrypted and can track the location of the users by realizing different attacks such as advanced inference attacks \cite{wasserkrug2008inference} and differential attacks \cite{wei2020ldp}.  

\paragraph{Data Usage and Consent}
Continuous data collection through multi-modal sensors including cameras, microphones, and other sensors will be closely involved in the daily routine activities of metaverse users. These sensors collect continuous data regardless of privacy awareness, which is logged over the local or global server(s). Consequently, this procedure raises legal questions regarding the users' consent and the kind of data that is collected and shared. Moreover, metaverse service providers can also utilize this data to optimize their inference models and to make them robust and more personalized. However, it also raises privacy concerns such as the collection, disclosure, and sharing of the data without the explicit or implicit consent of users.  

\paragraph{User Profiling} 
Similar to the current social media (in which users are considered as the product), everything will be a product alike in the metaverse. Metaverse will act as a meta-platform for different entities (such as users, developers, content creators, businesses, etc.), and this raises questions about data collection and its utilization for user profiling \cite{dimetaverse}. Also, the provisioning of the metaverse services requires that the users should be uniquely identifiable in the metaverse. For this purpose, VR headsets/glasses or any other such device can likely be used for illegally tracking users in real life \cite{shang2020arspy}. Moreover, such devices can be attacked by malicious actors and can be exploited to track users' real locations for possible digital and real-world crimes. Guzman et al. \cite{de2019security} presented a data-centric perspective to avoid unprecedented privacy challenges related to data collection and its usage in MR.

\subsubsection{Lack of Trustworthiness}
According to the definition of Trust, expressed by Lee and See \cite{lee2004trust}, in the perspective of automated systems, ``\textit{Trust is an attitude that an agent will help achieve an individual's goals in a situation characterized by uncertainty and vulnerability}.'' Metaverse is a data-driven technology in which service providers will use different automated tools to assist human beings as a recommendation engine in various domains, e.g., shopping recommendations, movie recommendations, and even recommendations regarding their health. The efficacy of such recommendation systems is highly dependent on the collection of personalized data for intelligent decision-making, e.g., an AI model is trained using the collected fine-grained health data to suggest more accurate recommendations regarding the well-being of users. Despite the huge potential of such features of AI-XR metaverse applications, it also raises many questions about which parameters have influenced the underlying AI model in producing a particular decision and the trustworthiness of such predictions and decisions. Next, we discuss such challenges across three dimensions (i) Truthful AI, (ii) Transparency, and (iii) Explainable and Interpretable AI.

\paragraph{Need for Truthful AI}
The evolution in technology also brings threats among them. Over the past few years, AI-based personal assistants including Siri, Alexa, Astro, and the like, have managed to gain wide social acceptance and these devices are being used by consumers daily for automating household routine tasks. Currently, misinformation, falsehood, or malfunction in AI-based text or speech analyses and command execution are not considered a matter of concern. However, it is expected that AI-enabled intelligent systems with linguistic capabilities will be a major feature of AI-XR metaverse applications. In this regard, it will be quite challenging to enforce the criterion of truthfulness in AI-based systems to ensure the safe selection of statements and behavior according to the social norms of users while interacting with human society.

\paragraph{Transparency Issues}
In general, practices, developing AI-based systems is about training sophisticated algorithms on large-scale data to learn an efficient and generalizable model that can be deployed in the real-world environment. However, it is well-known fact that the performance of these models is directly proportional to the transparency of training data, i.e., AI models will perform as well as their training data. Numerous factors such as input data riddled with poorly cleansed, or selection of inherently biased data, underfitting, and overfitting influence the performance of these models and can result in fairness and accountability issues (discussed later in this section). Unlike typical application development, there are no quality assurance tools available to spot bugs and evaluate the bias factor in the training data. For instance, if it was known at which stage, the model is going to infer at a perfect scale, then there would not be a need to perform training on such large-scale data. This process is all about the hit and trial procedure, which is a quite challenging task to identify the right approximations with better data, hyperparameters, and configuration settings.

\paragraph{Explainable and Interpretable AI}
The rapid adoption of based applications in human society has also grown the complexity of the systems, which ultimately requires system understandability to make them legitimate and trustworthy. In a critical human-facing technology such as the AI-XR metaverse, interpretable and explainable AI models are required to answer questions about accountability and transparency of their decisions and outcomes. For example, how the employed model reached the decision, and which factors influenced the models to make that decision~\cite{ali2022tamp}. Such questions are particularly important for human-centric applications where the potential impact of AI will be limited if it is not able to provide accurate and transparent AI predictions. Therefore, the key objective of these models is to develop a relationship of trust between human users and AI. However, one of the main challenges in developing explainable methods is the trade-off between achieving the modesty of an algorithm and ensuring the discretion of sensitive user data. In addition, it is also a challenging task to identify the right information while generating a simple yet useful explanation for users. It is worth noting that the terms interpretability and explainability are closely related and are often used interchangeably in machine learning literature, however, these terms are different in practice. Interpretability of the models is defined as the extent to which its outcomes are predictable, i.e., for a given change in the input or model parameter(s), the interpretability enables us to predict the respective change in its output. On the counter side, explainability deals with the explanation of internal processes of the ML/DL models in a human-understandable way.

\subsubsection{ML Security Issues}
Despite the state-of-the-art performance of modern AI techniques including DL-based systems, it has been shown that these models are highly vulnerable to carefully crafted adversarial perturbations known as adversarial ML attacks \cite{szegedy2013intriguing}. The threat of these attacks has been already demonstrated for many critical applications like healthcare, autonomous vehicles, etc. On a similar note, AI-XR metaverse applications are essentially critical as they involve humans, and ensuring their safety from any harm is profoundly important. On the counter side, the existence of these challenges raises many concerns about the safety, security, and robustness of AI-based metaverse applications thus hindering their practical deployment. As it is equally important that any AI-based should be equally trusted by all stakeholders involve service providers, developers, and end users. These challenges are detailed later (Section \ref{sec:SecML}).    

\subsubsection{Lack of Fairness and Accountability}
Modern AI methods like advanced DL models lack fairness and accountability in their decisions \cite{qayyum2020secure}. On the other hand, such questions are particularly important for critical applications like AI-XR, in which the model's decisions can have life-threatening consequences for the end users. Moreover, AI models are developed using training data, which will be mainly collected from human users in AI-XR metaverse applications for providing immersive experiences. Humans possess certain biases that will be readily reflected in the data they generate, and when this data is used for training AI models, the data bias will be directly translated into the developed AI-based system. As a result, the model will be biased towards certain samples that contain certain features (bias), and its decision will not be fair. On a similar note, the critical nature of AI-XR metaverse applications demands accountable decisions. Consequently, data bias if remained unaddressed can ultimately lead to unintended consequences \cite{latif2019caveat}. 

\subsubsection{Identity Theft and Authentication Attacks}
Users'/avatars' identities in the metaverse can be stolen or impersonated illegally leading to authentication and access control issues in the interconnected virtual worlds. Identity theft in the metaverse will be more dangerous than traditional attacks. The identity of a user once stolen will reveal everything about that person's digital assets, avatars, and social relationships. The attackers can exploit different vulnerable VR gadgets and other service authentication loopholes to realize identity theft attacks and can steal the victim's secret keys of digital assets and bank details. It has been reported that about 17 users in the OpenSea NFT marketplace were hacked through a phishing attack and flaws in the smart contract that resulted in a loss of \$1.7 million.\footnote{https://threatpost.com/nft-investors-lose-1-7m-in-opensea-phishing-attack/178558/}

Metaverse will leverage different biometrics and password-driven technologies to authenticate users and their avatars in the virtual worlds. The attacker can evade such authentication systems to impersonate real users' identities to get control of the whole virtual world. Evading AI-based biometric systems has become easier with the advancements in adversarial ML research. Therefore, AI-empowered speech and face recognition-based biometric systems can be easily attacked to realize impersonation attacks. Once the attacker has the access to the metaverse it can exploit the data generated by the victim's devices to deceive him, committing a crime in the virtual space. On other hand, the exposure of biological data when used for authentication purposes can also lead to severe consequences \cite{kurtunluouglu2022security}. Moreover, the authentication of social friends of a user using their avatars is much more challenging in the metaverse as compared to real-world identity authentication. In this regard, facial data, voice, and videos can be used to develop an AI-based avatar authentication system, however, the unsolved inherent issues of AI can still hinder its practicality.  

\subsubsection{The Bias Problem}
Bias refers to a model making certain unethical assumptions about the data. Human bias along with its many aspects has been studied by researchers in many disciplines including law, psychology, and so forth. In \cite{ntoutsi2020bias}, bias is defined as `the prejudice or inclination of a decision made by an AI system which is in a way considered to be unfair for or against one person or group'. Bias in recommendation systems, advertising algorithms, facial recognition systems, and risk assessment tools has been widely studied in recent years. In metaverse applications, data will be collected from a heterogeneous group of people and sources having their own characteristics, stereotypes, and behaviors, which introduces different biases in the collected data. In \cite{olteanu2019social, suresh2019framework, mehrabi2019survey}, the authors discuss different kinds of bias based on the sources and the types of bias. On the base of sources, these biases have been divided into further categories: biases caused by data, biases caused by algorithms, and biases caused by user interaction.

\subsection{XR-related Challenges in Metaverse}
AI-XR metaverse applications are essentially human-centric and ensuring the security, privacy, security, and robustness of such applications is of utmost importance. It has been envisioned that an entirely new form of digital media will emerge from the use of VR and AR in the modern metaverse (TV, print media, and the web). In recent years, there has been immense discussion regarding the concerns about surveillance capitalism, which is happening on the Internet in different applications. Many large tech organizations providing Internet services like Facebook, Google, Microsoft, and Amazon collect large of amount data related to the surveillance of their users, which is then used to satisfy the needs of advertisers \cite{vallor2016technology}. The pioneers of VR and AR such as Jaron Lanier \cite{lanier2018ten}and Louis Rosenberg\footnote{https://bigthink.com/the-future/metaverse-augmented-reality-danger/} have predicted that the concerns about surveillance are expected to rise in Metaverse. For example, it has been shown how reconfiguring AR in Pokémon Go (an AR mobile game) drew unexpected audiences to museums and public spaces like trains to fill in the space thus creating a form of virtual trespassing. It has been reported that people were putting their lives in danger to pursue virtual characters. This highlights that safety concerns may arise when such immersive technologies are engineered for gaming and experiences. Below we discuss the key challenges that are hindering metaverse applications in general and we will later discuss the specific challenges that arise with the use of different AI techniques in metaverse applications (Section \ref{sec:SecML}).  

\subsubsection{Safety Issues}
There are various concerns regarding the mental and physical safety of metaverse users. There are several reported incidents of digital harassment, theft, and bullying in XR applications \cite{Marr2021XR}. The report on ``Immersive and Addictive Technologies'' highlights rampant incidents of sexual harassment, cyberbullying, and grooming online \cite{uk2019immersive}. Ensuring the safety of users is a major challenge for AI-XR metaverse applications because of the fact that such incidents have real damage and harm to users despite being experienced in the virtual world. The avatars generated using recent advancements in AI techniques, in particular, generative models can appear more realistic in AI-XR metaverse applications and can engage users in promotional conversation thus providing a false sense of a real human behind the avatar. The avatars in such a promotional are fueled with more personalized data (such as your vitals, emotions, expressions, etc.) to look more realistic. Also, these sales avatars can pitch products to you more persuasively than any real salesman or even a recommendation system due to their access to rich cyber-physical data about you. The research in deep fake technology and photorealistic avatars is already on the stage where computer-generated content is indistinguishable from the real. Such advancements can be leveraged to realize an adversarial attack on AI-XR metaverse applications to get the intended behavior and outcomes. 

\subsubsection{Potential Antisocial Aspects}
There are various opinions regarding the antisocial aspects of AI-XR metaverse applications, many people think that introducing AI-XR metaverse may detract the users (humans) from their purposes and may have a somatic effect. In the literature, it has been shown that extended times online can result in users demonstrating post-VR sadness and detachment from reality. For instance, Aldous Huxley in 1932 wrote in his social science dystopian fiction novel that using technology can lead to self-inflicted harm that can lead people to be diverted from their higher priorities and become more prone to being influenced by other interests. As a result of such a quest for technological utopia, the human psyche and society as a whole are greatly afflicted. Social critics have long argued that various digital media, such as television, the Internet, and the Web, make people docile and less connected to the real world. For example, Jerry Mander in 1978 wrote in his book, ``Four Arguments for the Elimination of Television'' that TV removes the sense of reality from people, promotes capitalism, TV can be used as a scapegoat, and all these three factors work together negatively. The modern technological disruptions including the web and social networking services have created a filter bubble detaching people from the real world and the truth. Due to these reasons, the current era is also referred to as a post-truth era \cite{flintham2018falling}. We may reasonably expect that alienation from the real world will exacerbate with the increasing adoption of VR, AR, and AI-XR metaverse applications, which aim at changing the human perspective of the world. This argument can be supported by the fact that in 2018, the World Health Organization formally included ``gaming disorder" in its International Classification of Diseases following research that shows that technology can promote addictive behavior in people. Moreover, the literature focused on analyzing the social implications of metaverse argues to understand and identify potential psychological problems that can arise in metaverse \cite{buck2022security}. 

\subsubsection{Ethical Aspects}
Any technological intervention involving humans suffers from some serious ethical issues, especially the one that contains intelligence. The Institute of Electrical and Electronics Engineers (IEEE) has recently published a report on Ethically Aligned Design that mainly focuses on the Ethics of Autonomous and Intelligent Systems \cite{shahriari2017ieee}. This report emphasizes the need of developing ethical autonomous and intelligent systems (A/IS) that promotes human wellbeing and protects human rights through transparent and accountable A/IS and the prevention of the misuse of AI. This report is the collective effort of hundreds of researchers having diverse backgrounds and expertise in important areas like governance, technology, civil society, and policy-making. This report has a dedicated section on XR and interestingly, IEEE also has a Global Initiative on Ethics of XR.\footnote{https://standards.ieee.org/industry-connections/ethics-extended-reality/} In this report, various ethical issues related to XR have been highlighted including users' preference for virtual life over the real world and complete disengagement with society. In addition, the reports conclude with the following remark regarding XR: \textit{``The nature of XR environments fosters unique legal and ethical challenges that can directly affect users' privacy, identity, and rights. Society will need to rethink notions of privacy, accessibility, and governance across public and private spaces. New laws or regulations regarding data ownership, free use, universal access, and adaptive accessibility within XR environments may need to be developed.''}

\subsubsection{Regulatory Challenges}
The big tech companies have resisted regulation decrying the fact that regulation will slow down innovation. However, there are various ethics researchers and social scientists who are arguing for much greater regulation to ensure that consumer rights are protected. In this regard, the EU General Data Protection Regulation (GDPR) has paved the way for many countries and regions attempting to develop similar regulatory laws to protect Internet users. These regulations mainly emphasize the importance of the non-profiling of users (i.e., limiting the storage of tracking data) and for better transparency (i.e., online services and applications should specify why and what information is being stored). On a similar note, AI-XR metaverse applications are subject to the requirement of being transparent in terms of data collection and utilization and should also be subject to the informed consent of users. Also, the development of such applications requires thoughtful deliberation from a regulatory perspective. For instance. is worth considering banning non-medical applications to collect vital biomedical statistics due to the high risks of being exploited maliciously. To mitigate the risk of users being manipulated deceptively, metaverse operators may be bounded to transparently declare the staging of virtual products and experiences in the metaverse. Rosenberg, one of the pioneers of VR and AR, has already started to argue about the need for regulation for metaverse applications \cite{rosenbergregulation}. For instance, he suggested leveraging the arguments regarding the regulation of social media for developing a legal and philosophical basis for metaverse regulation. As the metaverse can be deemed as an evolutionary expansion of similar services. Rosenberg argued that the only solution to eliminate ethical and privacy-related concerns associated with metaverse is to shift from an advertising-based to a subscription-based business model in which users pay a subscription fee for accessing the metaverse platform. This eliminates the service providers' need to monitor their user base to a greater extent, however, this is not a feasible solution as it is difficult to say whether or not users will pay for a safer metaverse. 



\section{Analyzing Security and Trustworthiness Aspects of AI-XR}
\label{sec:SecML}
In this section, we will discuss the challenges associated with the use of different AI techniques (in particular, ML/DL-based models) that hinder the safe, secure, and trustworthy deployment of these methods in metaverse applications. We start by first providing a broad overview of AI security in the metaverse.  

\begin{figure*}[!t]
    \centering
    \includegraphics[width=0.99\textwidth]{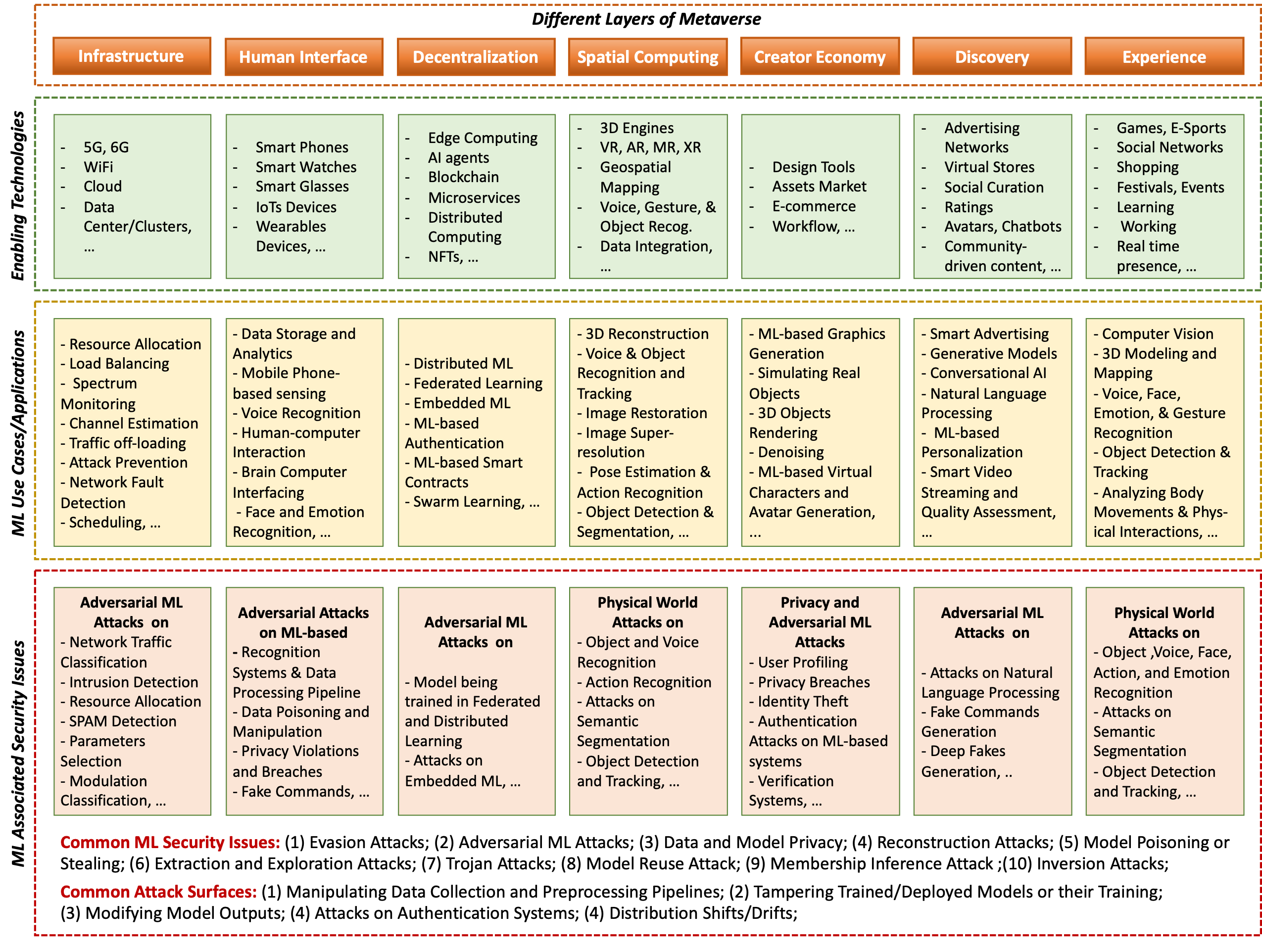} 
    \caption{Overview of ML security in Metaverse.}
    \label{fig:ML_Security}
\end{figure*}

\subsection{Security of ML in Metaverse}
The impact of metaverse applications will be social and economic and these applications will be more susceptible to undesirable adversarial action(s). AI will be the fundamental driving force behind the success of the metaverse, there are numerous applications of AI in different layers of the metaverse. On the other hand, the use of AI algorithms in AI-XR metaverse applications also opens them up to different adversarial attacks. In Figure \ref{fig:ML_Security}, we highlight the threat of different security and privacy attacks that can be realized in different applications in almost every layer of the metaverse. The figure also highlights that there are various common ML security issues and attack surfaces that get shared across the architectural landscape of the metaverse across different AI applications at each level. In this section, we discuss different AI-associated security and privacy attacks on AI-XR metaverse applications. 

\subsection{Potential Attacks on AI-based Metaverse Applications}
The threat of adversarial ML attacks has already been shown to be successful in compromising the integrity of AI techniques in many critical tasks, e.g., connected and autonomous vehicles \cite{qayyum2020securing}, computer vision \cite{akhtar2021advances}, and healthcare \cite{qayyum2020secure}, just to name a few. Furthermore, AI-XR algorithms could be biased either due to data imbalance or adversarial subversion. Many of the ethical dilemmas and social harms such as distraction, narcissism, disinformation, outrage, and polarization stem from the economic model of surveillance capitalism in which service providers give the customers everything and anything that makes the company money. In this way, these companies pander to the base animal desires of people and exploit their cognitive biases effectively downgrading humans and manipulating them for ulterior selfish purposes.

In the adversarial ML literature, an adversarial example is defined as the input to the deployed AI model crafted by an adversary by introducing imperceptible noise into the legitimate sample to get the intended outcomes. In general, there are two types of adversarial ML attacks: (1) poisoning attacks that aim at altering the training process of the AI model; and (2) evasion attacks that are focused on evading the deployed AI model by making inferences (they are also known as inference time attacks). In poisoning attacks, the adversary mainly modifies the training data to tamper with the learning of the AI model \cite{biggio2012poisoning}. In contrast, test data is manipulated in evasion attacks to get the desired predictions from the model \cite{biggio2013evasion}. Recent works have shown that AI models are vulnerable to attacks at both training and inference stages \cite{khalid2020fadec}. Training stage attacks typically corrupt a small subset (typically $\sim$1\%) of the training data samples to achieve malicious goals during AI model training~\cite{gu2019badnets, ali2020has}. On the other hand, the inference stage attacks cause a trained model to misbehave on adversarially crafted test inputs~\cite{ali2021all, ali2022detect}.

Attacks on AI models are generally carried out by first defining a threat model. A threat model is a set of assumptions regarding the attackers' abilities to access and affect a typical AI model training pipeline. Broadly, there are two main threat models---the poisoning threat model (i.e., realizing poisoning attacks), and the adversarial threat model (realizing evasion attacks). A poisoning threat model assumes an attacker who can control a small set of the training dataset to adversely affect the training of the model. An adversarial threat model assumes an attacker who can access and, to a certain extent, perturb the inputs to an already trained AI model. In the following, we highlight major security threats associated with the use of AI techniques in metaverse applications that include computer vision, natural language processing (MLP), network communication, authentication, and recognition systems. 



\subsubsection{ML Associated Security Issues in Computer Vision} 
Computer vision is one of the central building blocks in the foundation of the metaverse. In recent years, DL algorithms have enabled major advances in computer vision ranging from image classification \cite{deng2009imagenet, butt2021convolutional} to scene understanding \cite{cordts2016cityscapes, butt2022carl} and generating realistic images \cite{xia2021tedigan}. However, the discovery of the adversarial vulnerabilities of DL-based image processing models by Szegedy et al. \cite{szegedy2013intriguing} sparked a growing concern regarding the reliability and security of these deep models \cite{khalid2019qusecnets, ali2019sscnets, usama2018adversarial, usama2019adversarial}. Numerous works have analyzed these adversarial vulnerabilities in greater depth under different threat models~\cite{khalid2020fadec}. In general, adversarial attacks work by optimizing the perturbation, $\Delta x$, to an input image, $x$, such that the output of the model, $\mathcal{F}$, is significantly changed, $\maximize || \mathcal{F}(x) - \mathcal{F}(x+\Delta x) ||$. $\Delta x$ is typically optimized based on the gradients which are either computed directly (white-box scenarios) or estimated by introducing random noise (black-box scenarios). Summary of various adversarial ML attacks on different computer vision applications can be seen in Table \ref{tab:cv}. 

\begin{table*}[!ht]
\centering
\caption{Summary of different adversarial attacks on various computer vision applications (that are expected to be potential metaverse applications). }
\label{tab:cv}
\scriptsize
\resizebox{\linewidth}{!}{
\begin{tabular}{|l|c|p{55mm}|l|l|}
\hline
\multicolumn{1}{|c|}{\textbf{Application}} & \multicolumn{1}{|c|}{\textbf{Authors}} & \multicolumn{1}{c|}{\textbf{Methodology}} & \multicolumn{1}{c|}{\textbf{Datasets}} & \multicolumn{1}{c|}{\textbf{Before $\rightarrow$ After}} \\ \hline

\multirow{4}{*}{Face Authentication} & Goswami et al. \cite{goswami2018unravelling} & Studied how different architectures affect adversarial vulnerabilities. & MEDS, PaSC & 89.3\% $\rightarrow$ 41.6\% \\ \cline{2-5} 
& Sharif et al. \cite{sharif2016accessorize} & Developed adversarial glasses to fool face recognition systems. & Celebrity Face & 98.95\% $\rightarrow$ 0\% \\ \cline{2-5} 
& Shen et al. \cite{shen2019vla} & Developed black-box attack for face recognition systems using visible light. & CusFace, LFW & 100\% $\rightarrow$ 7.9\% \\ \cline{2-5} 
& Chatzikyriakidis et al. \cite{chatzikyriakidis2019adversarial} & Perturbed facial images to fool automatic face recognition to secure a person's identity. & CelebA & 97.8\% $\rightarrow$ 4\% \\ \cline{2-5} 

& Dabouei et al. \cite{dabouei2019fast} & Studied the vulnerability of face recognition systems against geometrically perturbed faces. & VGGFace2 & 100\% $\rightarrow$ 0.14\% \\ \cline{2-5} 
& Zhong et al. \cite{zhong2020towards} & Used dropout and feature-level attacks to improve the transferability of adversarial inputs. & VGGFace2 & 100\% $\rightarrow$ 3.24\% \\ \cline{2-5} 
& Dong et al. \cite{dong2019efficient} & Used evolutionary algorithm to find adversarial inputs against the models' decisions. & LFW & 100\% $\rightarrow$ 0\% \\ \cline{2-5} 
& Wenger et al. \cite{wenger2021backdoor} & Proposed improved physically-realizable attack against face recognition. & VGGFace & 100\% $\rightarrow$ 10\% \\ \cline{2-5} 
& Ali et al. \cite{ali2020has} & Proposed multi-trigger backdoor attack against backdoor defenses. & Celebrity Face & 88\% $\rightarrow$ 8\% \\ \cline{2-5} 
& Xue et al. \cite{xue2021backdoors} & Exploit hidden facial features as triggers of the backdoor attack. & VGGFace & 100\% $\rightarrow$ 0.02\% \\ \hline

\multirow{3}{*}{Object Detection} & Zhang et al. \cite{zhang2020contextual} & proposed generalizable contextual adversarial perturbations against object detectors. & PascalVOC, COCO & 78.8\% $\rightarrow$ 1.6\% \\ \cline{2-5} 
& Lee et al. \cite{lee2019physical} & Showed that non-overlapping physical patches can fool object detectors. & COCO & 55.4\% $\rightarrow$ 0.05\% \\ \cline{2-5} 
& Xie et al. \cite{xie2017adversarial} & Proposed multi-targeted adversarial attacks to fool object detectors. & PascalVOC & 72.07\% $\rightarrow$ 3.36\% \\ \cline{2-5} 

& Xie et al. \cite{xie2017adversarial} & Showed that multi-targeted adversarial attacks against object detectors are transferable. & PascalVOC & 54.87\% $\rightarrow$ 37.9\% \\ \cline{2-5} 
& Wei et al. \cite{wei2018transferable} & Utilized generative methods to efficiently obtain transferable adversarial inputs. & PascalVOC & 43\% $\rightarrow$ 3\% \\ \cline{2-5} 
& Wang et al. \cite{wang2020adversarial} & Utilized position and label information to attack black-box object detectors. & PascalVOC & 100\% $\rightarrow$ 16\% \\ \cline{2-5} 

& Wu et al. \cite{wu2022just} & Leveraged natural rotations to insert a backdoor into the object detectors. & PascalVOC & 89.5\% $\rightarrow$ 4.45\% \\ \hline

\multirow{5}{*}{3D-Object Modelling} & Wang et al. \cite{wang2021adversarial} & Optimally generates adversarial perturbations against 3D-Object detectors. & KITTI & 84\% $\rightarrow$ 0\% \\ \cline{2-5} 
& Xiang et al. \cite{xiang2019generating} & Generated 3D adversarial point clouds against Point-Net model. & ModelNet40 & 93\% $\rightarrow$ 0\% \\ \cline{2-5} 
& Hamdi et al. \cite{hamdi2020advpc} & Exploited an auto-encoder to generate transferable 3D adversarial perturbations to point cloud. & ModelNet40 & 93\% $\rightarrow$ 5\% \\ \cline{2-5} 
& Meloni et al. \cite{meloni2021messing} &  Used off-the-shelf 3D surrogates to transfer attack on 3D object models. & N/A & 100\% $\rightarrow$ 0\% \\ \cline{2-5} 
& Li et al. \cite{li2021pointba} & Proposed a novel formulation to develop backdoor triggers against 3D point cloud models. & ShapeNetPart & 98.4\% $\rightarrow$ 0.5\% \\ \hline

\multirow{5}{*}{Semantic Segmentation} & Arnab et al. \cite{arnab2018robustness} & Performed an in-depth study of adversarial vulnerabilities of semantic segmentation models. & Cityscapes & 77.1\% $\rightarrow$ 19.3\% \\ \cline{2-5} 
& Xie et al. \cite{xie2017adversarial} & Proposed multi-targeted adversarial attacks to fool semantic segmentation models. & PascalVOC & 72.07\% $\rightarrow$ 3.36\%\\ \cline{2-5} 
& Hendrik et al. \cite{hendrik2017universal} & Analyzed universal adversarial perturbation to fool a segmentation model for any input. & Citscapes & 64.8\% $\rightarrow$ 12.9\%  \\ \cline{2-5} 
& Li et al. \cite{li2021hidden} & Poisoned the segmentation models using object-level target class and semantic triggers. & ADE20K & 37.7\% $\rightarrow$ 25.2\%  \\ \cline{2-5} 
& Feng et al. \cite{feng2022fiba} & Proposed frequency-injection backdoor attack against medical image segmentation tasks. & KiTS-19 & 54.5\% $\rightarrow$ 21.1\% \\ \hline
\end{tabular}
}
\end{table*}

\subsubsection{ML Associated Security Issues in NLP}
Similar to the vulnerabilities of ML models for vision applications, the literature demonstrates that the ML methods for modeling NLP tasks are also vulnerable to malicious attacks, at both the training and the inference stages of a typical ML pipeline~\cite{ali2020has, ali2021all}. Below we discuss such attacks. 

\noindent \textit{Poisoning and Trojaning Attacks:} Poisoning attacks and trojaning (also known as backdoor) attacks are the most widely known training stage attacks in NLP. Poisoning attacks aim to tamper with the training of the model so that it is unable to perform satisfactorily on the test inputs~\cite{biggio2012poisoning, munoz2017towards, steinhardt2017certified}. Trojaning attacks aim to insert a trojan---typically characterized by a specific pattern of words known only to the attacker in the input sequence---into a model such that the model behaves normally on natural test inputs, but malfunctions as desired by the attacker (through the use of of the trojan pattern of words~\cite{gu2019badnets, shafahi2018poison}). Similar to the case with the computer vision applications, the trojaning attacks, being more difficult to be detected as compared to the poisoning attacks, pose a greater threat to NLP applications in the metaverse.

\begin{table*}[!ht]
\centering
\caption{Summary of different adversarial ML attacks on various NLP applications. }
\label{tab:nlp}
\scalebox{0.83}{
\begin{tabular}{|l|c|p{85mm}|l|l|}
\hline
\multicolumn{1}{|c|}{\textbf{Application}} & \multicolumn{1}{|c|}{\textbf{Authors}} & \multicolumn{1}{c|}{\textbf{Methodology}} & \multicolumn{1}{c|}{\textbf{Datasets}} & \multicolumn{1}{c|}{\textbf{Before $\rightarrow$ After}} \\ \hline





\multirow{2}{*}{Language to Language Modelling}  & Zhand et al. \cite{zhang2021crafting} & propose word saliency speedup local search method to attack translation machines. & NIST (MT) & 92.48\% Degradation \\ \cline{2-5}

& Boucher et al. \cite{boucher2022bad} & Uses invisible characters, homoglyphs and deletion control characters to fool the model & WMT14 & 37\% $\rightarrow$ 1\% \\ \hline

\multirow{9}{*}{Fake-news Detection} & Li et al. \cite{li2020bert} & exploit BERT-MLM to fool a fine-tuned BERT model by generating coherent perturbations. & AG news & 94.2\% $\rightarrow$ 10.6\% \\ \cline{2-5}
& Ali et al. \cite{ali2022detect} & propose an adaptive adversarial attack to generate perturbations against statistical defenses. & Kaggle Fake news & 95\% $\rightarrow$ 0\% \\ \cline{2-5}
& Jin et al. \cite{jin2020bert} & identify key contributing words and replace them with synonyms while retaining the coherence. & Kaggle Fake news & 96.7\% $\rightarrow$ 15.9\% \\ \cline{2-5}

& Zellers et al. \cite{zellers2019defending} & Present a generative model, Grover, to generate fake-news that fools fake-news detectors. & Not applicable & 95\% $\rightarrow$ 67\% \\ \cline{2-5} 

& Pan et al. \cite{pan2022hidden} & Exploits linguistic styles as triggers to backdoor an NLP model. & COVID & 95.1\% $\rightarrow$ 6.7\% \\ \hline

\multirow{8}{*}{Toxicity/Sentiment Classification} & Garg et al. \cite{garg2020bae} & exploit BERT-MLM to generate adversarial perturbations that are coherent with the context. & Amazon & 96\% $\rightarrow$ 11\% \\ \cline{2-5}
& Boucher et al. \cite{boucher2022bad} & Uses invisible characters, homoglyphs and deletion control characters to fool the model. & Wikipedia Detox & 95\% $\rightarrow$ 19.5\% \\ \cline{2-5}
& Ebrahimi et al. \cite{ebrahimi2017hotflip} & Leverage atomic flip operation to swap tokens to fool NLP fake news classifiers. & AG news & 92.35\% $\rightarrow$ 27.7\% \\ \cline{2-5} 
& Li et al. \cite{li2018textbugger} & Exploit model gradients to find and perturb the most positively contributing words. & IMDB & 90.7\% $\rightarrow$ 0\% \\ \cline{2-5} 

& Li et al. \cite{li2020bert} & exploit BERT-MLM to fool a fine-tuned BERT model by generating coherent perturbations. & IMDB & 90.9\% $\rightarrow$ 11.4\% \\ \cline{2-5} 
& Jin et al. \cite{jin2020bert} & identify key contributing words and replace them with synonyms while retaining the coherence. & Yelp & 93.8\% $\rightarrow$ 1.1\% \\ \cline{2-5} 

& Chen et al. \cite{chen2021badnl} & Use char- and word-level triggers to backdoor NLP sentiment classifiers. & SST-5 & 55\% $\rightarrow$ 0\% \\ \cline{2-5} 
& Irtiza et al. \cite{irtiza2022sentmod} & propose a context-aware hidden trigger backdoor attack against NLP classifiers. & IMDB & 84.5\% $\rightarrow$ 2.48\% \\ \hline

\end{tabular}
}
\end{table*}

\noindent \textit{Adversarial Attacks.} Although adversarial examples have been extensively studied in computer vision, they have received significantly limited attention in NLP tasks mainly due to the discrete input search space---minimal adversarial perturbations in the input are no longer feasible in NLP~\cite{ali2021all, ali2022tamp}. Recently, however, there have been numerous works highlighting the adversarial vulnerabilities of the NLP-based ML models. Notable adversarial attacks include Text-bugger, Text-fooler, PWWS, and BERT Adversarial Example~(BAE).

Adversarial attacks against NLP models generally follow three major steps---evaluation, perturbation, and selection---to achieve some adversarial goal---for example, targeted or untargeted misclassification---under a predefined threat model. Consider, for example, an input sequence $X = \{x_1, x_2, ..., x_i, ...,x_n\}$ correctly classified by an NLP model, $\mathcal{F}$, in class, $\mathcal{F}(X)=y \in \mathbb{R}^M$. At the evaluation stage, the attacker uses some \textit{impact scoring function}, to compute a set, $I_x$, representing the impact of each word over the output. At the perturbation stage, the attacker repeatedly perturbs the most impactful words in $I_x$ using some pre-defined \textit{perturbation mechanism} such that the semantic and contextual value of $X$ remains preserved. At the selection stage, the most optimal perturbation is selected. Table \ref{tab:nlp} provides a summary of various adversarial ML attacks on different NLP applications.

\subsubsection{ML Associated Security Issues in Networking}
AI-XR metaverse applications will provide ubiquitous connectivity to a massive number of users over wireless networks. Over the last few years, many AI-based algorithms have been developed to improve the performance of wireless communication and networking systems that will be used in different layers of network architecture \cite{chen2019artificial}. The use of AI in wireless communication empowers wireless devices to perform many important intelligent functions such as network composition, analyzing traffic patterns, managing content requests, analyzing wireless channel dynamics, etc. Moreover, AI-based algorithms have been used for optimizing different network constraints like high throughput and low latency for different multimedia applications. A prominent use case is to leverage intelligent proactive load management in 5G and 6G communication networks and predictive data analytics to improve network operations. Despite the significant potential of using various AI algorithms for different optimizing applications in wireless networks, recent studies have highlighted that AI application is highly susceptible to adversarial ML attacks. For instance, Usama et al. \cite{usama2019generative} used a generative adversarial network (GAN) for realizing adversarial attacks on network intrusion detection. The threat of adversarial ML attacks on network traffic classification is demonstrated in \cite{usama2019black} and for cognitive self-driving networks is presented in \cite{usama2018adversarial,usama2019adversarial}. Similarly, the threat of adversarial ML for 5G networks is analyzed in  \cite{usama2021examining}. Summary of various adversarial ML attacks on network applications is presented in Table \ref{tab:network}. We refer interested readers to a detailed survey highlighting the threat of adversarial ML in network security \cite{ibitoye2019threat}.

\begin{table*}[!ht]
\centering
\scriptsize
\caption{Summary of different adversarial ML attacks on different networking applications. }
\label{tab:network}
\resizebox{0.98\linewidth}{!}{
\begin{tabular}{|l|c|p{55mm}|l|l|}
\hline
\multicolumn{1}{|c|}{\textbf{Application}} & \multicolumn{1}{|c|}{\textbf{Authors}} & \multicolumn{1}{c|}{\textbf{Methodology}} & \multicolumn{1}{c|}{\textbf{Datasets}} & \multicolumn{1}{c|}{\textbf{Before $\rightarrow$ After}} \\ \hline

\multirow{2}{*}{Intrusion Detection} & Usama et al. \cite{usama2019generative} & Exploited GAN to craft adversarial examples to evade intrusion detection model. &  KDD99 & 89.12\% $\rightarrow$ 56.55\% \\ \cline{2-5}
& Aiken et al. \cite{aiken2019investigating} & Perturbed a few features to evade four ML classifiers trained for detecting DDoS attacks.  & KDD99 & 100\% $\rightarrow$ 0\% \\ \hline

\multirow{1}{*}{Network Traffic Classification} & Usama et al. \cite{usama2019black}  & Crafted adversarial examples using mutual information in black-box settings. & UNB-CIC Tor Data & 96\% $\rightarrow$ 77\% \\ \hline

\multirow{1}{*}{Modulation Classification} & Usama et al. \cite{usama2021examining}  & Used C\&W attack to evade traffic modulation classifier.  & RML2016.10a & 85\% $\rightarrow$ 15\% \\ \cline{2-5}
& Sadeghi et al. \cite{sadeghi2018adversarial} & Realized white-box and black-box attacks on VT-CNN model using a PCA-based perturbations. & GNU Radio & 75\% $\rightarrow$ 38\% \\ \hline

\multirow{1}{*}{Network Modulation} & Usama et al. \cite{usama2021examining}  & Realized black-box attack on channel autoencoder on unsupervised and DRL models. & RML2016.10a & 95\% $\rightarrow$ 80\% \\ \hline 

\multirow{1}{*}{Malware Classification} & Usama et al. \cite{usama2018adversarial}  & Realized three SOTA adversarial ML attacks, i.e., FGSM, BIM, and JSMA. & Malware Image Data \cite{usama2019adversarial} & 98.39\% $\rightarrow$ 1.87\% \\ \hline 

\multirow{1}{*}{Abnormal KPI Detection} & Usama et al. \cite{usama2019adversarial} & Leveraged two SOTA attacks to evade ML-based abnormal KPI detection classifiers. &  LTE network data & 98.8\% $\rightarrow$ 13.7\% \\ \hline 

\multirow{1}{*}{Channel State Estimation} & Sagduyu et al. \cite{sagduyu2019iot} & Realized three attacks: spectrum poisoning, jamming, and priority violation. & Not Articulated & 95.58\% $\rightarrow$ 23.12\% \\ \hline 
\end{tabular}}
\end{table*}


In addition to the above motioned adversarial vulnerabilities associated with the use of AI techniques in many network applications, some other critical network-related issues can hinder the smooth operation of metaverse at a global level. For instance, centralized network architecture provides flexibility in terms of cost saving, simplicity, and ease in performing different operations. On the other hand, such architectures are more prone to a single point of failure (SPoF) and distributed denial of service (DDoS) attacks \cite{wang2021blockchain}. For example, if a powerful attacker gets control of the network, it may lead to severe challenges like SPoF and DDoS. To address such issues, the literature suggests leveraging decentralized network architecture \cite{nguyen2021metachain}. In addition, decentralization will potentially amplify the transparency and trust of users in exchanging their virtual belongings (like digital assets and virtual currencies) among each other and across different virtual worlds in the metaverse. However, many issues arise with the use of decentralized approaches, e.g., reaching a consensus on an ambiguous operation among the huge number of entities in a dynamic metaverse. \textit{Distributed Denial of Service (DDoS):} Metaverse will include a massive number of IoT devices, which can be compromised by an attacker to form a botnet to realize DDoS attacks \cite{bertino2017botnets}. \textit{Sybil Attacks:} In a Sybil attack, the adversary pretends to have fake (or manipulated) identities of legitimate users or devices. Using such stolen identities he can take over the network. 


\subsubsection{Security Issues in Cloud-hosted ML Models}
\begin{figure*}[htp]
    \centerline{\includegraphics[width=0.98\textwidth]{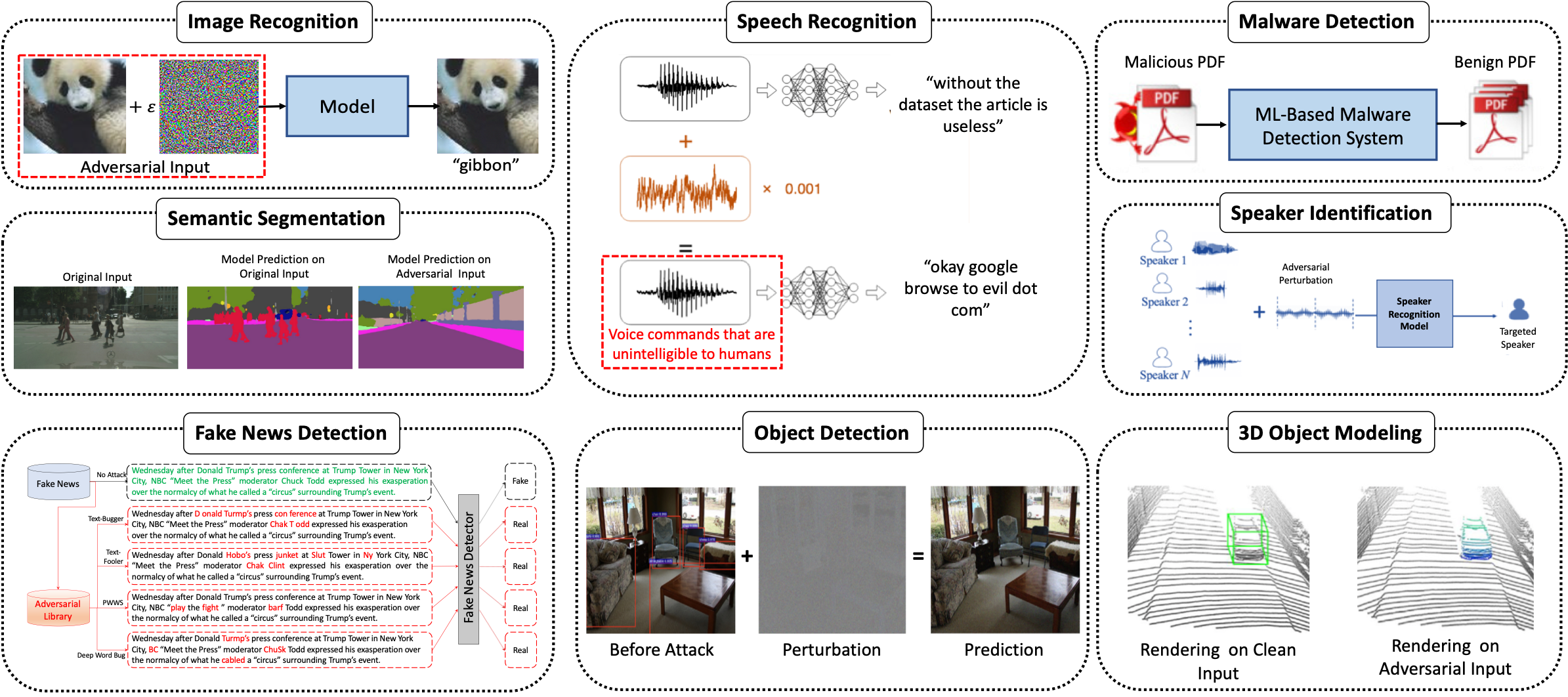}}
    \caption{Illustration of adversarial ML attacks on different potential applications in AI-XR metaverse. Individual figure references: Image Recognition \cite{szegedy2013intriguing}; Speech Recognition \cite{qin2019imperceptible}; Malware Detection \cite{xu2016automatically}; Semantic Segmentation \cite{fischer2017adversarial}; Speaker Identification \cite{xie2020real}; Fake News Detection \cite{ali2021all}; Object Detection \cite{wang2020adversarial}; and 3D Object Modeling \cite{wang2021adversarial}.}
    \label{fig:visuals}
\end{figure*}

Outsourcing the training of ML/DL models to third-party services that offer powerful computational resources on the cloud is prevalent nowadays. These services allow ML developers to upload their data and models for training over their cloud platforms. It is expected that such services will be featured in AI-XR metaverse applications, as they provide the flexibility of developing AI models using sufficiently large training datasets while reducing the cost and time. However, the literature demonstrates that such services are vulnerable to variety of attacks such as backdoor attacks \cite{chen2020backdoor}, exploration attacks \cite{sethi2018data}, model inversion \cite{yang2019neural} and model extraction attacks \cite{kesarwani2018model}, etc. More details about various attacks and defenses for cloud-hosted ML models can be found in \cite{qayyum2020securing_bd}. Visual illustration of adversarial ML attacks on different potential applications in the AI-XR metaverse is presented in Figure \ref{fig:visuals}. 

\subsection{Attacks on VR}
The literature highlights that VR systems are vulnerable to adversarial attacks. For instance, Casey et al. \cite{casey2019immersive} demonstrated that humans in VR systems can be controlled like joysticks--thus providing the adversary the ability to control the movement of VR user without his consent or getting into his knowledge. Moreover, the literature suggests that both security and privacy attacks can be realized on VR/XR systems \cite{valluripally2020attack}. Therefore, developing secure and robust AI-XR metaverse systems is crucial to the widespread adoption of metaverse applications that are not vulnerable to adversarial attacks or are capable to withstand such attacks and mitigating their impact.

\subsection{Analyzing Implications of ML Security, Privacy, and Trust Issues: An AI-XR Case Study}
\begin{figure*}[htp]
    \centerline{\includegraphics[width=0.95\textwidth]{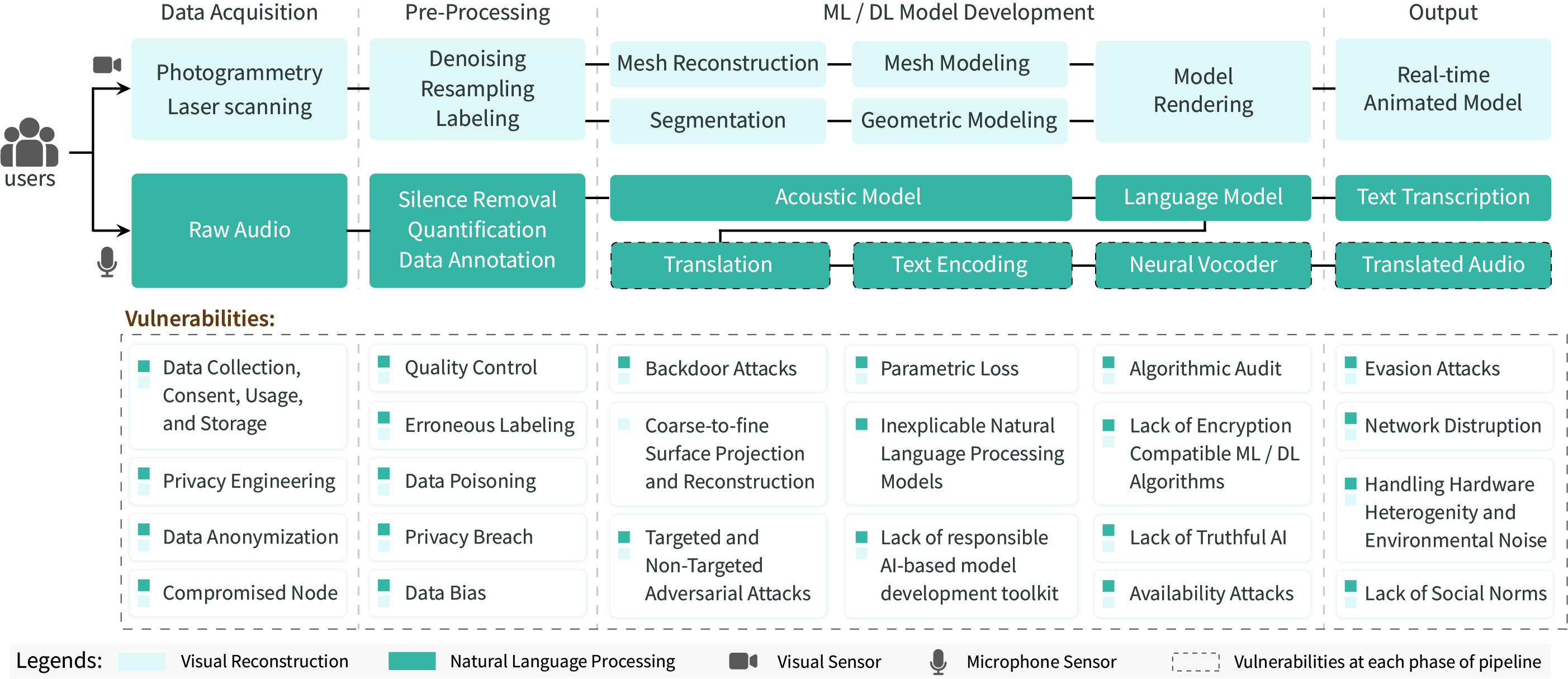}}
    \caption{A prospective \textbf{pipeline for developing AI-XR metaverse applications} for multi-lingual communications, involving various security challenges at each stage.}
    \label{fig:pipeline}
\end{figure*}

In this section, we present an ML/DL-based pipeline for a potential AI-XR metaverse application use case. We then analyzed various challenges and threats that can arise at each development stage. The pipeline is developed while considering a general metaverse application---a virtual conference, in which the participants are remotely connected from different places (the pipeline is presented in Figure \ref{fig:pipeline}). A unique avatar is representing each participant while each avatar is expected to reflect real-time voice, facial expressions, and gestures. The voice of each participant is translated into the native language of all the participants along with generating the transcription. The pipeline depicts different ML/DL-empowered tasks: (1) \textit{3D/4D Visual Reconstruction}---responsible for generating photo-realistic avatars; (2) \textit{3D Visual Mapping}---to reflect real-time multi-modal expressions (i.e., audio, facial, and gestures, etc.); (3) \textit{Speech Recognition and Synthesis}---to interpret and translate the voice of recipient into other languages; and (4) \textit{Speech-to-Text Synthesis}---to generate the transcription of audio conversations of all members. 

As depicted in Figure \ref{fig:pipeline}, data acquisition is performed by collecting raw audio input through a microphone for NLP, whereas, depth cameras and laser scanners are used for 3D/4D visual reconstruction and mapping. In the next step, acquired data is pre-processed through several techniques including data denoising, deblurring, silence removal, etc. The processed data is then labeled for the training of ML/DL models in a supervised/semi-supervised learning fashion. After successful data preparation, 3D visual mesh construction and segmentation models are trained to perform 3D avatar reconstruction. On the other hand, acoustic and language models along with neural vocoders are trained to perform the multilingual translation and transcription tasks. Although in the literature, these pipelines have demonstrated significant performance in 3D reconstruction and NLP tasks, however, this pipeline is highly exposed to the various privacy and security attacks at each stage of the development pipeline (as shown in Figure \ref{fig:pipeline}).

\section{Towards Developing Secure and Trustworthy AI-XR}
\label{sec:solutions}
\begin{figure*}[!ht]
    \centering
    \includegraphics[width=0.95\linewidth]{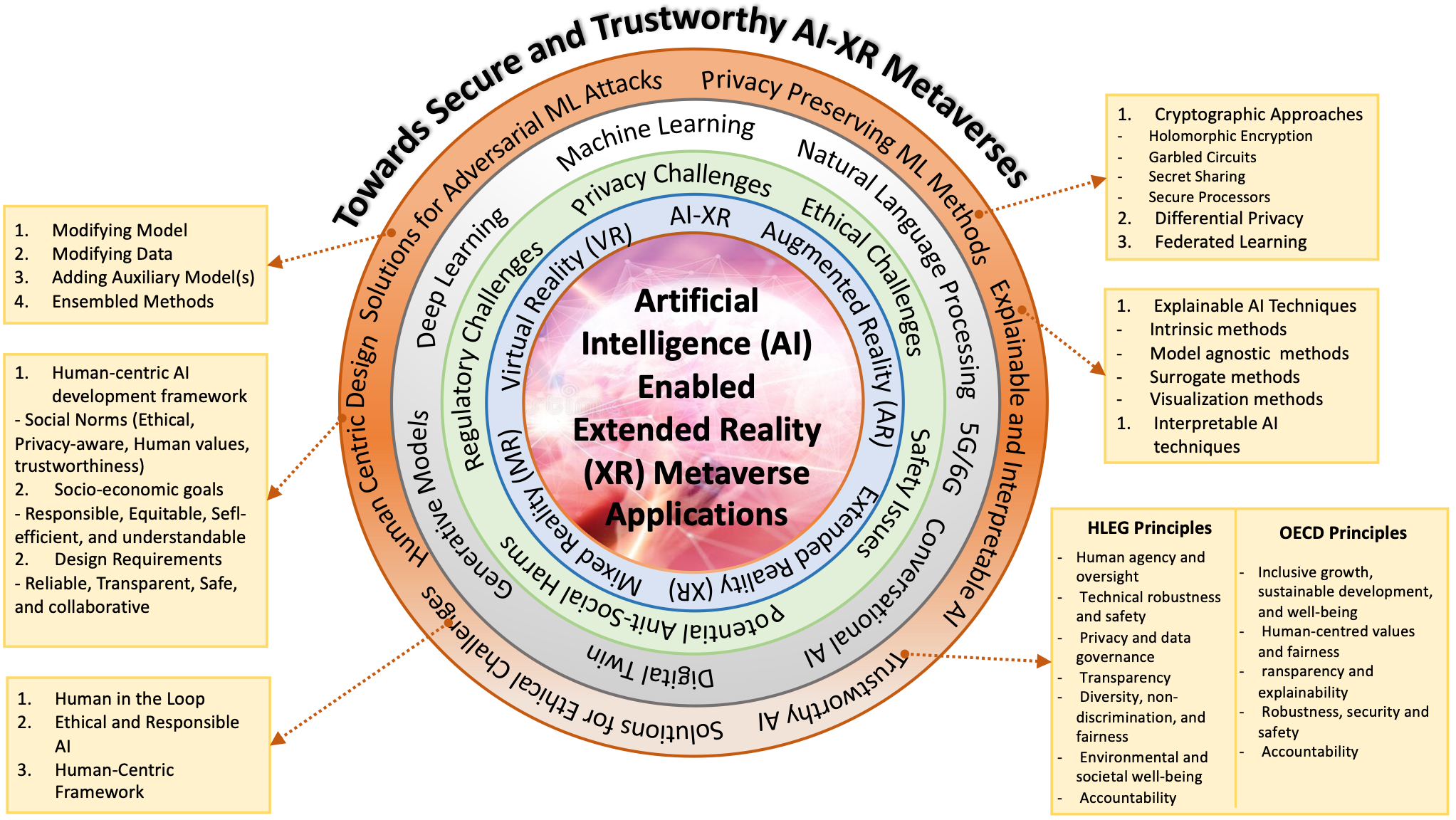} 
    \caption{An abstraction of different ML-associated challenges along with a taxonomy of various solutions that can be used to address those challenges.}
    \label{fig:solutions}
\end{figure*}

The development of secure, safe, and trustworthy AI-XR metaverse applications is fundamentally very important, in this section we will discuss different potential solutions that can be leveraged to address challenges associated with the use of AI in particular and for the overall system in general. An abstraction of different techniques that can be leveraged to address the ML-associated issues is shown in Figure \ref{fig:solutions} and these methods are described next. 

\subsection{Solutions for Privacy Protection in AI-XR}
In the literature, privacy-preserving techniques are broadly categorized into three classes: (i) cryptographic techniques, (ii) differential privacy, and (iii) federated and distributed ML. These techniques are briefly discussed below.

\subsubsection{Cryptographic Techniques}
Cryptography refers to a practice of methodologies, aiming to construct and analyze communication protocols to ensure secure communication while achieving data integrity, authentication, non-repudiation, and data confidentiality. Generally, there are two common types of encryption methods: (i) symmetric encryption, and (ii) asymmetric encryption method. The symmetric encryption method is a secret-key algorithm, in which the sender and receiver must share the same key to perform encryption and decryption of the data. Whereas, asymmetric encryption method (also known as public-key cryptography) uses two keys, i.e., public and private key, associated with an entity which require to authenticate its identity electronically or encrypt data. The public key of each entity is published whereas, the corresponding private key is always kept secret to perform encryption or decryption of data. In literature, Ron Rivest, Adi Shamir and Leonard Adleman (RSA) \cite{rivest1978method}, Data Encryption Standard (DES) \cite{standard1999data}, Advanced Encryption Standard (AES) \cite{pub2001197}, and Secure Hash Algorithm (SHA) \cite{burrows1995secure} are a few commonly used algorithms used for data encryption. Different cryptographic techniques can be employed to convert readable information to an encrypted state, which can be later used at the receiver end after performing decryption. Below we discuss some of the most commonly used encryption methods that can be used for the development of privacy-aware AI models.



\paragraph{Homomorphic Encryption} Homomorphic encryption (HE) is a computational approach that performs encryption while allowing computational tasks to be executed over encrypted data at the same time to ensure the privacy of the data. HE is defined as a public key cryptographic technique in which a pair of public and private keys is created to perform encryption and decryption operations on the data. The public key is used to encrypt the data, before sharing it with the third party for further computational tasks including training, and/or inference. Due to the homomorphic characteristics of this approach, the results can be decoded using the private key to visualize the results without showing them to third-party servers or unauthenticated users. In the ML literature, HE has been used for protecting the privacy of the users' data for different applications such as genome imputation \cite{sarkar2021fast}, misinformation detection in text messages \cite{alispam}, etc. Specifically, the AI models are trained and inferred using encrypted training and testing data thus preserving the privacy of the sensitive data.  

\paragraph{Secure Multi-party Computation} Secure multi-party computation (also known as secure computation) is a type of cryptographic technique that is focused on the development of collaborative methods to perform joint computation and calculate a function over joint inputs while possessing those inputs in an isolated fashion. Contrary to the traditional cryptographic methods, where cryptography ensures confidentiality and integrity of communication or storage, while the adversary is outside the system of users, this approach protects the users' privacy from each other while performing ML-based tasks including training and inference activities.

\paragraph{Garbled Circuits} The idea of garbled circuits was first proposed by Yao in 1986 to perform two-party computation \cite{yao1986generate}. Garbled circuits can be used in a scenario where multiple parties are interested in performing some computation without sharing their data. Let's assume two parties (e.g., Alice and Bob for the sake of simplicity) want to perform some computation using garbled circuits. Alice will send his input and function in the form of a garbled circuit and Bob will utilize his garbled input with the garbled circuit to get the result of the required function, once he obtains it from Alice in an oblivious fashion. In \cite{bost2015machine}, garbled circuits along with HE have been used to develop privacy-aware ML models, where the authors trained three classification models namely decision tree, Naïve Bayes, and hyperplane decision classifier using the encrypted data.  

\paragraph{Secret Sharing} In secret sharing, multiple parties collaborate in the computation by sharing their secrets among them while holding a ``share'' of the individual secrets. The secret can only be reconstructed by combing all the individual shares kept by participating parties, otherwise, it will be useless. In the literature, the secret sharing technique has been successfully used for training AI models in a privacy-preserving way. For instance, Bonawitz et al. \cite{bonawitz2017practical} used the secret sharing technique to train an ML model by aggregating model updates from multiple parties in a privacy-aware way. In a similar study \cite{hossain2019emotion}, authors used this technique for the development of a privacy-aware ML-based emotion recognition system leveraging client-server architecture. In their proposed framework, the secret sharing technique was used for the communication of audio-visual data from the client side to the server, where an ensemble model based on a sparse autoencoder and a CNN model was used for the feature extraction from the collected data. The SVM classifier was then trained using the extracted features for the emotion recognition task. A secret sharing-based parallelized variant of principal component analysis (PCA) for preserving data privacy is presented in \cite{bogdanov2018implementation}. 

\paragraph{Secure Processors} Secure processors were pioneered by rogue software to protect sensitive code from being accessed by malicious actors at higher privilege levels. Secure processors are being used in different processors now to perform privacy-preserving operations, e.g., the Intel SGX processor. In \cite{ohrimenko2016oblivious}, SGX processors were used to developing a data oblivious system for different ML techniques that include SVM, decision tree, matrix factorization, and k-mean clustering. The primary goal was to facilitate collaboration between multiple data proprietors performing the ML task on an SGX-empowered data center.

\subsubsection{Differential Privacy}
The idea of differential privacy is based on introducing noise in the data to protect sensitive information while ensuring the usefulness of the data after noise addition \cite{dwork2011differential}. Differential privacy is defined in terms of the task-specific concept of neighbor datasets and it provides strong guarantees in ensuring the privacy of the data during algorithmic analysis \cite{abadi2016deep}. Numerous differential privacy-based methods have been presented in the literature, such as differentially-private stochastic gradient descent (DP-SGD) \cite{du2021dynamic}, private-aggregation of teacher ensembles (PATE) \cite{zhang2020towards}, exponential noise based differential privacy-preserving methods to ensure privacy on large-scale data. These methods demonstrated better applicability in ML-based applications in various domains including intelligent transportation services, smart/virtual personal assistants, and smart healthcare services.

\subsubsection{Federated Learning}
Federated learning (FL) refers to a distributed-ML paradigm that is capable of learning global ML models without directly accessing and/or exchanging data from edge devices. Intuitively, basic FL-based methods consist of a collaborative learning framework where each participant such as an edge device, network node, and local server can independently train a model using its local data. These edge devices then share their model parameters with a server, which then performs aggregation of the parameters after receiving parameter updates from each edge device. Finally, the server updates the parameters of the global model and shares the updated parameters with all participants. The iterative process of FL is continuous until the desired criteria as been fulfilled, e.g., validation accuracy/loss or the maximum number of communication rounds. In this way, a global model is trained without requiring the actual data from the FL participants. Subsequently, this sharing mechanism allows ML-based systems to learn from large-scale diverse data and develop a global model. Such methods can demonstrate better applicability in terms of dealing with sensitive data in various human-centered applications such as AI-XR metaverse applications. Despite the success of FL in training an ML model with reliable performance while maintaining the privacy of the actual data, different attacks can be realized on the model being trained using the FL paradigm, e.g., backdoor attacks \cite{xie2019dba}, label flipping attacks \cite{qayyum2022making}, free-riding attacks \cite{lin2019free}, and poisoning attacks \cite{ali2021incentive}, etc. Also, it has been demonstrated that sensitive information can be extracted from the shared parameters in FL settings \cite{boenisch2021curious}. 

\subsection{Solutions to Combat Adversarial ML Attacks in AI-XR}
In the literature, adversarially robust ML models have been mainly categorized into three categories \cite{qayyum2020secure}: (1) Data Modification; (2) Model Modification; and (3) Using Auxiliary Model. Moreover, a few methods leverage a hybrid approach in which multiple defensive techniques are used to develop adversarially robust ML models. Below we discuss the most prominent methods in each category and we refer the interested readers for more details about these methods to recent and comprehensive surveys that are specifically focused on adversarial ML \cite{qayyum2020securing,akhtar2021advances,wang2022survey,yuan2019adversarial}.  

\subsubsection{Data modification} Data modification methods work by modifying the input data during the training or inference phase to mitigate the effects of adversarial perturbation. A few famous data modification methods are briefly described below. 

\begin{itemize}
    \item \textit{Adversarial Re-training:} This method was proposed by Goodfellow et al. \cite{goodfellow2014explaining} and it is considered to be a basic method for mitigating the effect of adversarial perturbation in the trained model. In this method, adversarial examples are augmented in the training data, which is then used to (re)-train the model. This method has been extensively used in the literature, however, a few research studies demonstrated that the models trained using this method are not robust against multiple attacks \cite{tramer2019adversarial}.
    \item \textit{Feature Squeezing:} Xu et al. \cite{xu2017feature} presented a feature squeezing-based approach that aims to squeeze feature space of input that may be exploited in response to an adversary. In this regard, the heterogeneous feature vectors have been collectively joined into a single space to reduce available feature space. Although, the proposed defense method achieved significant performance against small perturbations. However, it was found less effective against iterative adversarial attacks \cite{he2017adversarial}.
    \item \textit{Input Reconstruction:} Input reconstruction-based defense methods have been proposed to mitigate the effect of adversarial attacks. These methods transform adversarial examples into legitimate samples by cleaning adversarial noise using an appropriate technique, e.g., using an autoencoder to clean adversarial perturbations \cite{gu2014towards}.  
\end{itemize}

\subsubsection{Model modification} Model modification methods aim at modifying the parameters of trained ML models to defuse the effect of adversarial attacks. The most commonly used model modification methods are described below. 

\begin{itemize}
    \item \textit{Gradient Regularization:} This method allows complex neural networks to bring a partial surge in training computational complexity to improve the performance of the network regardless of any prior knowledge about adversarial attacks. This idea was coined by Ross et al. \cite{ross2018improving} to improve the performance of CNN models on classification tasks. Though the proposed method achieved significant improvement in CNNs' robustness, it also increases the computational cost of models which prejudices the performance in real-world ML-based applications.
    \item \textit{Defensive Distillation:} Distillation in a neural network was initially conceptualized by Hinton et al. \cite{hinton2015distilling} to establish knowledge sharing from a larger network to a smaller one. Later, Papernot et al. \cite{papernot2016distillation} extended this notion by developing a distillation-based defense mechanism against adversarial attacks, which is known as defensive distillation. In this method, the larger model is trained over hard labels to maximize accuracy while predicting the output probabilities of the baseline smaller model. This method is successful in mitigating the effect of small adversarial perturbation and it fails in the presence of strong adversarial perturbations, e.g., adversarial examples generated using C\&W attack \cite{carlini2017adversarial}.
    \item \textit{Network Verification:} In this method, certain properties of the ML/DL model are verified, e.g., validating the output of models, produced in response to the corresponding input samples. Katz et al. \cite{katz2017reluplex} presented ReLU and satisfiability modulo theory (SMT) based network verification method to make complex neural networks robust against adversarial examples. In a similar study, authors have proposed a scalable quantitative verification framework for DNNs to prove formal probabilistic property against adversarial attacks \cite{9402111}.
\end{itemize}

\subsubsection{Using Auxiliary Model} Methods aiming to robustify ML models in this category use an additional model either for detection of adversarial examples or for clean adversarial perturbations. A few methods are described below.

\begin{itemize}
    \item \textit{Adversarial Detection:} In such methods, a detector model is used to differentiate between normal and adversarial inputs, e.g., a binary classifier \cite{lu2017safetynet}. 
    \item \textit{Ensembling Defenses:} In this defense strategy, an ensemble of different defensive techniques is created to withstand different adversarial attacks. PixelDefend is the most famous ensemble defense method that consists of two defense approaches, i.e., input reconstruction and adversarial detection \cite{song2017pixeldefend}. 
    \item \textit{Using Generative Modeling:} These types of methods leverage different ML/DL-based generative models for cleaning adversarial noise in adversarial examples to project them back to the same data manifold.   
\end{itemize}

\subsection{Solutions for AI-XR Transparency and Trust Challenges}
The true potential of AI-based applications in AI-XR metaverse applications can only be realized when they are developed using fine-grained personal data for making personalized recommendations and predictions, which is only possible when users fully trust the underlying system. Therefore, addressing the challenges related to the trustworthiness aspects of AI-XR metaverse applications is very important. From an AI perspective trustworthiness itself requires predictability, interpretability, explainability, safety, and robustness. Below we discuss different methods that can be used to accomplish trustworthiness in AI applications. 

\subsubsection{Explainable and Interpretable AI}
An AI model is referred to as explainable if it can explain the ability of parameters to justify the results. Explainability makes the AI models transparent which ultimately helps in evaluating and understanding the results provided by the models. In recent years, substantial research efforts have been conducted to enhance explainability, trustworthiness, and interpretability in AI models. Fairness, Accountability, and Transparency in Machine Learning (FAT-ML) \cite{fat2018fairness} and Defense Advanced Research Projects Agency (DARPA), explainable AI program \cite{gunning2017explainable} are the two famous research groups working in this context. The literature argues that explainable models can be the first step toward converting black-box AI models into white-box models \cite{adadi2018peeking}. 

Interpretable models refer to the models that explain themselves. In simple words, an AI model is said to be interpretable, if its decision against some input is logically understandable such as which factors influenced the AI model to reach that decision. In the literature, various methods have been presented to leverage interpretability in ML models. These methods ensure that the predictions of interpretable models are unbiased, which ultimately makes it easier to trust these systems in human society. It is worth noting that the terms interpretable and explainable are interchangeably used in the literature, however, they are different in terms of domain-specific definitions, moreover, there is no exact definition of these terms \cite{rasheed2022explainable}.  A detailed taxonomy of different explainable and interpretable AI methods can be found in \cite{adadi2018peeking,rasheed2022explainable}. 

\subsubsection{Trustworthy AI}
The relevant literature emphasizes two famous sets of principles that can be used to attain trustworthy AI. One of them is developed by European Commission's AI High-Level Expert Group (HLEG) \cite{holzinger2022information} and the other one is defined by Organisation for Economic Co-operation and Development (OECD) \cite{yeung2020recommendation}. The following are the seven essential principles outlined in OECD: (1) Human agency and oversight; (2) Technical robustness and safety; (3) Privacy and data governance; (4) Transparency; (5) Diversity, non-discrimination, and fairness; (6) Environmental and societal well-being; and (7) Accountability.


Similarly, the following principles are outlined in HLEG: (1) Inclusive growth, sustainable development, and well-being; (2) Human-centred values and fairness; (3) Transparency and explainability; (4) Robustness, security, and safety; and (5) Accountability. 
One of the key noticeable insights from the above two principles set is that they mainly emphasized explainability, security, fairness, safety, and robustness aspects of AI. Therefore, these are the essential requirements that need to be fulfilled to develop trustworthy AI-based applications. In addition, we can see that these principles are essentially human-centric that respect ethical norms. As potential AI-XR metaverse applications will be more human-focused, therefore, the above-mentioned principles can be leveraged to develop trustworthy AI-based applications for the metaverse. 

\subsection{Solutions for Ethical Challenges in AI-XR}

\subsubsection{Human in the Loop}
The metaverse's inherent complexity raises different security issues. For instance, it can be envisioned that the metaverse administrators will have to push automation, that is, to handle more tasks with algorithms, rather than with human operators, due to the requirement of managing a large number of users, applications, and services. The generated data will be much larger than those managed by the current Web platforms. Delegating tasks to algorithms, especially those implemented with state-of-the-art AI approaches is necessary to meet high-level efficiency and scalability. However, in the current version of social media and the Internet, we have even started to realize the implications of using algorithms for managing societally relevant tasks. Despite the significant performance, these algorithms suffer from various issues. Some authors writing on the governance of metaverse have proposed the use of a modular approach for the development of metaverse applications, as it allows adapting regulations to specific scenarios and then controlling the system accordingly \cite{fernandez2022life}.  

\subsubsection{Ethical and Responsible AI}
To ensure socially desirable AI decisions, novel ways are required to be figured out to simultaneously minimize potential harms associated with the use of AI and its potential benefits. In this regard, the importance of taking an ethics-first approach towards the development of AI-based technologies becomes more plausible \cite{floridi2021ethics}. However, there are many challenges associated with the development of ethical AI pipelines due to distinct social norms and demographics of the human population, i.e., one ethical solution may be beneficial for a group of people but it is highly possible that it will not be suitable for another group on the same time. Therefore, customized solutions are required to address such issues that can consider the social norms of target users while making AI-based decisions. In this regard, different ethical guidelines can be leveraged that can be potentially used for the development of pro-social AI solutions. The literature shows a groundswell of interest in ensuring ethical and responsible AI \cite{qadir2022toward}.


\subsection{Situational Awareness}
Situational awareness can be defined as the capacity to understand information perceived from the surrounding environment. The literature argues that situational awareness is a crucial and effective tool for monitoring the security of complex systems like metaverse \cite{woodward2022analytic}. Situational awareness can be used at the local and global levels for threat monitoring in a single metaverse or across multiple metaverses, respectively. The feasibility and potential of this tool have been extensively studied in the literature focused on XR and VR technology. For instance, Woodward et al. \cite{woodward2022analytic} performed a literature review that focused on the design of information presentation in AR headsets to enhance users' situational awareness. Authors in \cite{ju2020acoustic}, performed immersive and realistic simulations to evaluate the effectiveness of audio-visual warning systems in increasing users' situational awareness in accident situations using VR. They demonstrated that VR can assist drivers to remain alert in emergency situations.

\subsection{Human Centric Approach for AI-XR Development}
Metaverse is essentially a human-centric application \cite{heller2016avatars}. To realize the real social impact of different AI-XR metaverse applications, they should be analyzed and developed using human-centric design thinking. Metaverse service providers and developers must pay attention to key stakeholders (i.e., humans) by prioritizing and considering their social norms, i.e., dignity, justice, and rights, and supporting goals including creativity, self-efficacy, social connections, and responsibility. 
The aforementioned characteristics can be inherited in AI-XR metaverse applications by following three key concepts proposed in \cite{shneiderman2020human}. The first one is \textit{Human-centric framework}---that guides the developers and researchers to ensure human-centric thinking about high-level two-dimensional control. Secondly, \textit{Design metaphors}---which points out how two key goals of AI and social norms are both valuable. However, the stakeholders such as developers, researchers, policymakers, and business leaders must combine them both in developing metaverse applications to provide ultimate benefits to the users. Thirdly, \textit{Governance Structures}---which ensures the bridge between the above-mentioned ethical principles and the practical measures needed to achieve the desired goals including reliable metaverse application development while ensuring cultural safety to increase privacy and trustworthiness of the users.

\section{Open Research Issues}
\label{sec:open}
In this section, we highlight various open research issues that are particularly associated with the use of ML/DL models in different AI-XR metaverse applications. 

\subsection{Developing Generalizable Adversarial Defense Methods}
Over the past few years, substantial research attention has been devoted to adversarial ML. However, the literature highlights that the attention devoted to developing adversarially robust ML/DL models is significantly less as compared to developing novel attack methodologies \cite{qayyum2020securing_bd}. In the literature, different defensive techniques have been proposed to withstand adversarial ML attacks (as discussed above), however, each method only works in a specific setting and fails to withstand unseen and powerful attacks (consequently, fails to generalize across a wider class of attacks). On the other hand, the literature focused on adversarial ML shows that the diversity and severity of these attacks are increasing with each passing day. Therefore, the development of hybrid and universal defensive techniques is the need of the hour. In addition, it is required that the defense techniques should be developed while considering evolvable and adaptable adversaries (who can adapt their capabilities to break defense strategy). The threat of adversarial ML can be a major hurdle in the development of secure, safe, robust, and trustworthy AI-XR metaverse applications and if it remained unaddressed, can cause unintended severe consequences to users and society. It is highly recommended to consider these aspects while developing ML/DL-empowered human-centric applications like the AI-XR metaverse. Moreover, the worst-case robustness test can be performed from an adversarial lens considering different attack surfaces in individual AI-XR metaverse application architecture.  

\subsection{Investigating Robustness of Privacy-Preserving Methods}
As discussed above, AI-XR metaverse applications will collect fine-grained data that may include personal attributes to provide personalized services (empowered by ML/DL models). The models trained with such data can be inferred to reconstruct privacy-related information that can be exploited to get intended outcomes and incentives. Although different privacy-preserving ML techniques have been proposed in the literature that has been shown quite successful in preserving data privacy, however, the literature demonstrates that meaningful information can still be inferred even if the presence of an appropriate privacy-preserving method. For example, it has been demonstrated that homomorphic encryption (one of the widely used encryption techniques) is vulnerable to model extraction attacks \cite{reith2019efficiently}. Similarly, Boenisch et al. \cite{boenisch2021curious} showed that sensitive information can be reconstructed from the shared parameters in FL. This suggests that the investigation of vulnerabilities and limitations of existing privacy-preserving methods can be a good step toward developing robust privacy-preserving methods. Ideally, it is required that the ML/DL models should be developed in such a way that they are by design privacy-aware, i.e., they should not be able to learn any privacy-related features from the data that could be compromised upon model inferences. 

\subsection{Developing Generalizable Explainable and Intrepratble Techniques}
Another major limitation of DL models hindering their trustworthy applications in critical applications like the AI-XR metaverse is the lack of explainability and interpretability. This can also be exploited by adversarial agents to craft adversarial perturbations to realize attacks on different AI-XR metaverse applications. Although significant research interest has been devoted to the development of novel techniques to explain and interpret DL models, the literature shows that their application is limited to a certain data type or application \cite{rasheed2022explainable}. While the AI-XR metaverse applications will have a complex architecture that will simultaneously use multi-modal data for making different intelligent decisions, existing explainability and interpretability techniques cannot be directly used for explaining and interpreting ML/DL-driven decisions. More work is required to create methods that can be generalized across different data types, models, and applications.  

\subsection{Developing Ethical Data Analysis Pipelines}
Current ML/DL models are not capable of considering different ethical norms that are necessary for human-centric applications like AI-XR metaverse applications. On contrary, these considerations are yet very important to maximize potential benefits and minimize associated harms to ensure safe, robust, and fair data analysis. In AI-XR metaverse applications, different ML/DL models will be trained using massively large data collected by humans and their interactions with the real and virtual universe. While there is no guarantee that the AI decision will be ethically-committed because the data used for model training might contain data bias that will eventually result in biased decisions. Moreover, the outcomes of ML/DL models will be just a reflection of human behavior (including moral failures even if they are not intentionally committed). Therefore, to increase the trust of different stakeholders involved in AI-XR metaverse applications (particularly, end users) and to provide them with a sense of safety, fairness, and accountability, it is highly desirable to develop novel techniques to ensure fair and ethical data analysis empowered by various ML/DL techniques.  

\subsection{Pushing AI on Edge: Embedded ML}
One of the feasible approaches to preserve the privacy of AI-XR metaverse users will be to deploy ML/DL models on their smart devices, e.g., smartphones, AR/VR gadgets, tablets, etc. By doing so models can be developed and inferred on their devices without requiring to transmit data to a central cloud. We envision that various AI-XR metaverse applications will potentially adopt embedded ML or edge-enabled ML due to the proliferation of different enabling gadgets and smart devices. However, numerous challenges related to underlying hardware computing capabilities will arise when sufficiently large ML/DL models will be deployed on resource-constrained devices. Also, the literature argues that the research on enabling edge AI is at its early stages of development \cite{qayyum2022collaborative}. Therefore, it is worth investigating the feasibility and potential of deploying M/DL models on embedded devices to ultimately develop secure, private, and robust systems to provide personalized services in AI-XR metaverse applications. We refer interested readers to a recent survey on analyzing the notion of edge-enabled metaverse applications for a more comprehensive discussion on the topic and various challenges \cite{xu2022full}.  

\subsection{AI-XR Metaverse Specific Security Solutions}
The future AI-XR metaverse will have a complex structure and will be a combination of various enabling (complex) technologies (that possess their associated challenges related to privacy and security, e.g., adversarial ML). Moreover, the massive connectivity of numerous entities (users, services providers, organizations, etc.) along with the decentralization will even worsen the enormity of security and privacy in AI-XR metaverse applications. Individual vulnerabilities associated with each technology can be exploited to realize a more powerful attack to halt or get control of some services or the entire metaverse. If such vulnerabilities are left unaddressed, they will eventually lead to novel challenges thus making it challenging to ensure the secure, safe, and robust operation of metaverse services. Therefore, it is very crucial to understand such challenges and develop customized defense solutions to protect AI-XR metaverse applications and services in general.

\section{Conclusions}
\label{sec:con}
In this paper, we have analyzed various security, privacy, and trustworthiness challenges associated with the use of different machine learning (ML) and deep learning (DL) techniques in artificial intelligence and extended reality (AI-XR) metaverse applications. Specifically, considering the layered architecture of the metaverse, we developed a pipeline and highlighted different potential ML/DL use cases along with identifying various vulnerabilities associated with their application. Furthermore, we provide a comprehensive overview of these challenges and discuss potential solutions that could be used to overcome such issues. To accentuate the implications of adversarial threats, we designed a customized case study (considering a prospective AI-XR metaverse application) and analyzed its security and privacy aspects. Finally, we discussed various open research issues that require further investigation. We envision that our work on this crucial topic will provide a one-stop solution to interested researchers who aim to develop secure, robust, and trustworthy AI-XR applications.  

\bibliographystyle{IEEEtran}

\end{document}